\documentclass{article}
\usepackage{ amsmath, graphicx}

\usepackage[bookmarks=false,hyperfigures=false, hidelinks, breaklinks]{hyperref}
\usepackage[hyphenbreaks]{breakurl}

\usepackage{doi}
\usepackage{graphicx}
\pdfoutput=1

\usepackage[preprint]{neurips_2022}

\usepackage{verbatim}
\usepackage{ifpdf}
\usepackage{xspace}
\usepackage{multirow}
\usepackage{float}
\usepackage{caption, epstopdf}

\usepackage{amsmath, amsfonts, amsthm, bbm} % Math packages
\usepackage{algorithm2e}
\usepackage{amsmath}
\usepackage{algorithmic} %format of the algorithm

\usepackage{amssymb}

\usepackage{bm}

\usepackage{xcolor}
\usepackage{color}
\usepackage{setspace}
\usepackage{tabu}
\ifpdf
  \usepackage{graphicx}
  \DeclareGraphicsExtensions{.pdf,.eps,.jpg}
\else
  \usepackage{graphicx}
  \DeclareGraphicsExtensions{.pstex,.ps,.eps,.jpg}
\fi
\usepackage{color}
\usepackage{subfigure}

\setcitestyle{numbers}
\floatstyle{plain}
\newfloat{algorithm}{tbp}{loa}
\floatname{algorithm}{Algorithm}

\newcommand{\edit}[1]{%
{\color{black} #1}%
}%

\newcommand{\ours}{\texttt{MTNeuro}~}
\allowdisplaybreaks[2]

\author{
Jorge Quesada$^1$\thanks{ Equal contribution.~ Contact authors:  (ELD, JQ, LS) \{evadyer, jpacora3, lsathidevi3\}@gatech.edu; (ECJ) erik.c.johnson@jhuapl.edu; $\dagger$ Both senior authors contributed equally.} , Lakshmi Sathidevi$^{1*}$, Ran Liu$^1$, Nauman Ahad$^1$,  Joy M.~Jackson$^1$,\\ {\bf Mehdi Azabou$^1$, Jingyun Xiao$^1$,  Christopher Liding$^1$, Matthew Jin$^1$, Carolina Urzay$^1$},\\ {\bf William Gray-Roncal$^2$, Erik C.~Johnson$^{2,\dagger}$, Eva L.~Dyer$^{1,\dagger}$}\\
1- Georgia Institute of Technology \\
2 - Johns Hopkins University Applied Physics Laboratory %\\ \texttt{evadyer@gatech.edu}
}
\title{MTNeuro:  A Benchmark for Evaluating Representations of Brain Structure Across\\ Multiple Levels of Abstraction}

\begin{document}

\maketitle
\begin{abstract}
There are multiple scales of abstraction from which we can describe the same image, depending on whether we are focusing on fine-grained details or a more global attribute of the image. In brain mapping, learning to automatically parse  images to build representations of both small-scale features (e.g., the presence of cells or blood vessels) and  global properties of an image (e.g., which brain region the image comes from) is a crucial and open challenge. However, most existing datasets and benchmarks for neuroanatomy consider only a single downstream task at a time. To bridge this gap, we introduce a new dataset, annotations, and multiple downstream tasks that provide diverse ways to readout information about brain structure and architecture from the same image. Our multi-task neuroimaging benchmark (\ours\hspace{-1mm}) is built on volumetric, micrometer-resolution X-ray microtomography images spanning a large thalamocortical section of mouse brain, encompassing multiple cortical and subcortical regions. 
We generated a number of different prediction challenges and evaluated several supervised and self-supervised models for brain-region prediction and pixel-level semantic segmentation of microstructures.  Our experiments not only highlight the rich heterogeneity of this dataset, but also provide insights into how self-supervised approaches can be used to learn representations that  capture multiple attributes of a single image and perform well on a variety of downstream tasks.  
Datasets, code, and pre-trained baseline models are provided at: \url{https://mtneuro.github.io/}.

\end{abstract}

\section{Introduction}

Our understanding of our natural surroundings requires multiple levels of perceptual processing: we can recognize a macroscopic object (e.g., a tree), while also identifying finer-grain structures within it (e.g., leaves and branches), and context-relevant features (e.g., leafiness, height, or season). This multi-level perception scheme also translates to the medical image domain: the process of interrogating medical images (either by a human expert or an algorithm) involves combining macrostuctural insights (such as a region of interest) with context-relevant microstructure information 
and human-interpretable features (e.g., the density of a given cell type in a microscopy image) in order to derive a diagnosis or characterize a target sample.

In particular, the  ongoing effort to understand the connections and dynamics of the brain involves analyzing both macroscopic level properties such as region-level structures \cite{bassett2017small,liu2020generative} as well as detailed microstructures like the size or shape of a given cell type \cite{schneider2020chandelier}. While significant advances have been achieved in unveiling the properties and structures within the brain through several imaging modalities at different scales \cite{behrens2012human,hawrylycz2014allen,shapson2021connectomic, scala2021phenotypic}, most existing neuroimaging benchmarks are designed for evaluation at a single spatial scale, or geared towards a particular downstream task. This can be attributed to several causes, including the prohibitive cost of manual annotation for data spanning multiple scales \cite{lichtman2014big,motta2019big}, the associated computational cost of processing multi-scale data, and the fact that neuroimaging technologies have only recently progressed towards pipelines  that  can  capture  multi-area volumes at high resolutions. 

To fill this gap, we present the \ours benchmark (Figure~\ref{fig:overview}): a multi-task, multi-scale benchmark based on a large 3D X-ray microtomography image dataset spanning multiple areas from a mouse brain. Code and access to the data are provided at: \url{https://mtneuro.github.io/}. We host our dataset in the Brain Observatory Storage Service and Database (BossDB, a specialized interactive database \cite{hider2022brain}) and provide an integrated dataloader to facilitate transfer experiments and analysis. This benchmark provides a unified framework that allows for evaluating models and representations arising in three distinct tasks: 

\begin{itemize}
    \item {\bf Task 1 - Image-level classification of brain region:} prediction of the brain region (somatosensory cortex, striatum, thalamus, zona incerta) to which a given image belongs.
     
 \item {\bf Task 2 - Pixel-level segmentation of microstructures}: prediction of neural microstructures (axons, cell bodies, blood vessels, background) at the pixel-level across the four core brain regions in the dataset.
  
 \item {\bf Task 3 - Probing multiple semantic features from learned image-level embeddings}: estimation of semantic (human-interpretable) features (such as the average cell size or axon density) from the representation of a given image, obtained after ``freezing" the weights of a trained encoder. 
 
\end{itemize}

To understand how current models perform on these different tasks, we evaluate a family of different supervised and self-supervised models. Our results in Tasks 1 and 3 highlight a significant generalization gap between self-supervised and supervised approaches, which opens up interesting opportunities for further evaluation and development of self-supervised learning (SSL) methods for these tasks. Through testing across a family of different models across a variety of tasks, our proposed benchmark provides both an exciting platform for evaluating self-supervised learning (SSL) methods, and a rich tool in the effort to extract fundamental insights into brain architecture at both the micro- and macro-scale. %different levels of the brain.

\begin{figure*}[t!]
\centering
\includegraphics[width=\textwidth]{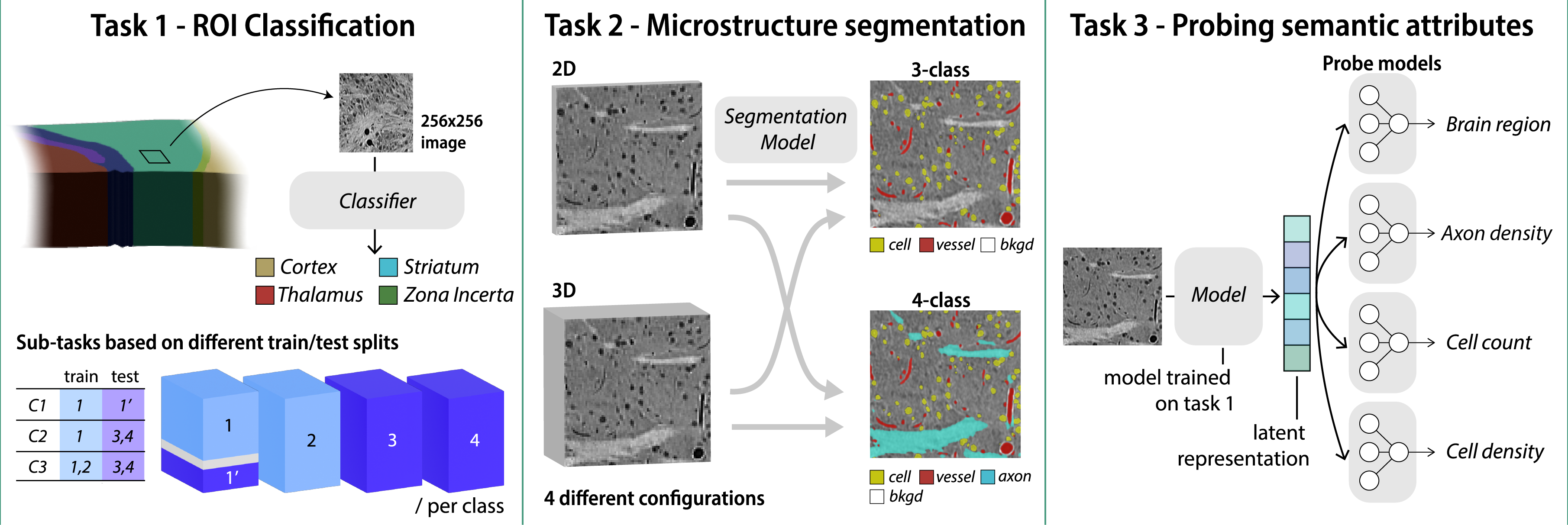}
\caption{\footnotesize {\em Overview of \ours Benchmark.} Task 1: Brain region (macrostructure) classification (3 configurations which vary in data availability and testing schemes); Task 2: Pixel-level microstructure segmentation (4 configurations which vary in sample dimensionality, span and target class count); Task 3: Probing semantic attributes from image-level embeddings obtained in Task 1.} \vspace{-3mm}
\label{fig:tasks}
\end{figure*}

\section{Background and Related Work}
\label{sec:background}

\subsection{The need for a benchmark in brain mapping and connectomics}
%Increasing resolution of techniques and scale- 
Over the past decade, there have been major advances in our ability to resolve fine-scale neuroanatomical structures in the brain. With these advances, we have generated large amounts of brain data that span many spatial scales, and can reveal different features of brain organization. At the nanoscale, electron microscopy has provided detailed wiring diagrams of small portions of cortex \cite{kasthuri2015saturated}. At micron scale, microscopy techniques have provided detailed pictures of cytoarchitecture - or how neurons and cells are organized \cite{hawrylycz2014allen}. Efforts at even larger scales to capture many brain areas simultaneously, like connectivity atlas and X-ray microtomographic datasets \cite{dyer2017quantifying}, have provided information about the interplay between long-range connections across brain areas and microstructures, such as cell body densities and other morphological features of brain structure. 

%Need for AI
Accompanying these new tools for data generation have been major advances in machine learning and computational approaches for modeling and analyzing these datasets, for problems such as object detection, segmentation, and classification. While the information provided by these methods are incredibly rich and have a great deal of structure at many scales, any given method is typically tested on an individual challenge at a particular scale. The expense of annotating and proofreading can be considerable, and significant neuroanatomy knowledge is typically required of annotators \cite{motta2019big}. Moreover, many efforts to provide high quality data, such as \cite{hawrylycz2014allen}, have not focused on building ML-oriented benchmarks, but rather on providing reference datasets and resources. 

%Unique challenge- represent macro and micro structure from same imaging
Using machine learning tools to understand these emerging brain datasets at different spatial scales is both a challenge and an increasingly critical need. As a result, large tera- and peta-scale connectomics datasets are being collected using electron microscopy and X-ray microtomography, including data from the entire brain of \emph{Drosophila} \cite{xu2020connectome,zheng2018complete}, large portions of the mouse brain \cite{prasad2020three, schneider2020chandelier} and even a cubic millimeter of human cortex \cite{shapson2021connectomic}. Advances in imaging technologies promise to continually increase the spatial extent, number of species, and number of imaged individuals. These datasets have the micro- or nanoscale resolution and large spatial extents required to resolve sub-cellular structures (e.g., mitochondria and synapses), microstructures (e.g., glia, neurons, and vasculature), and macrostructure (e.g., brain regions, cortical layer structure, and long-range white matter projections). The multi-scale nature and large size of these datasets requires new ML tools \cite{balwani2021multi}, which drives the need for benchmarks  that can extract representations of neural structure at different scales.

\subsection{Existing  datasets and benchmarks for resolving brain structure}
\label{sec:existing-datasets}
Due to the large variety of spatial scales, neuroanatomical structure, and imaging modalities, a wide range of segmentation and classification problems have been formulated for neuroimaging data. At the macroscale, there has been a long history of developing benchmarks for different datasets in MRI and related modalities like DTI and fMRI. For example, the BraTS dataset \cite{menze2014multimodal} focuses on the MRI of brain tumor and motivated many segmentation works \cite{chen2019self, Battalapalli2022Optimal, Amin2022New, Latif2022Glioma}.
The ADNI \cite{Jack2008ADNI} and MIRIAD \cite{malone2013miriad} datasets provide MRI-based Alzheimer’s disease imaging that focuses on tracking disease progression \cite{Liu2018MultiModalityCC, LU201826, islam2018brain, Liu2019joint, PUENTECASTRO2020103764, Marzban2020Alzheimer, DEEPA2022103455}. \edit{The UK Biobank \cite{UKBioBank2015} offers a huge collection of Brain MRI data from 5,00,000 human participants}. The low spatial resolution of MRI and related methods, however, limit the ability to observe microscale structures at the cellular or subcellular level. 

%However, one of the biggest limitation of MRI imaging technique is that the resolution is too low and thus makes it impossible to study the finer-scale microstructures inside the brain.

For microscale structure, there are fewer benchmarking efforts due to the scale and complexity of the annotation and processing. Example problems include segmentation of synapses in the CREMI challenge \cite{cremi}, which provides high-resolution imaging of synapses in 5 cubic microns of non-isotropic EM with nanometer resolution. Other problem formulations include 2D image segmentation of cell membranes (ISBI 2012 Challenge), with data from \cite{cardona2010integrated}, and 3D segmentation of cells \cite{lee2017superhuman,plaza2016focused}. Specific benchmarks have also been developed for axon instance segmentation \cite{wei2021axonem} and mitochondria segmentation \cite{wei2020mitoem}. Different forms of microscopy with micrometer resolution, including calcium fluorescence microscopy, are suitable for segmenting cell bodies and extracting functional time-series data, but lack other microstructure information. Benchmarks have also been established for estimation of functional traces from two-photon calcium fluorescence microscopy, including spikefinder \cite{berens2018community}. These benchmarks have been instrumental in driving progress on specific problems at specific spatial scales. In general, however, most benchmarks at the microscale focus on relatively small spatial extents, and lack the multi-scale macrostructure, which is also present in the brain. \edit{To the best of our knowledge, there are no public microtomography datasets of brain structure with both dense microstructure and macroscale annotations currently available.} 

Encompassing larger spatial extents, projects such as the BigBrain atlas provides high-resolution sections from the mouse brain with Nissl contrast \cite{amunts2013bigbrain}, but the resolution only allows resolution of cells around 20 \textmu m. Large-scale EM datasets are also being used to benchmark performance and segment neurons, augmented with iterative human proofreading \cite{januszewski2018high,li2019automated,shapson2021connectomic}, which are leading to large segmented datasets with increasingly complex annotation. These data, however, are not suitable for large-scale benchmarking and algorithm development in the general machine learning community due to their size and ongoing refinement. To accelerate progress towards machine learning tools which can operate at multiple levels of spatial abstraction within the same high-resolution dataset, benchmarks are needed which encompass large spatial extents at high resolution. This will enable a broader scientific community to apply state-of-the-art methods for these important applications.
\section{Dataset and Tasks}
\label{sec:method}
\subsection{Overview of dataset}
\begin{figure*}[t!]
    %\centering
    \begin{minipage}[b]{0.2\textwidth}
        %\centering
        \includegraphics[width=\textwidth, trim={10cm 2cm 10cm 0cm}]{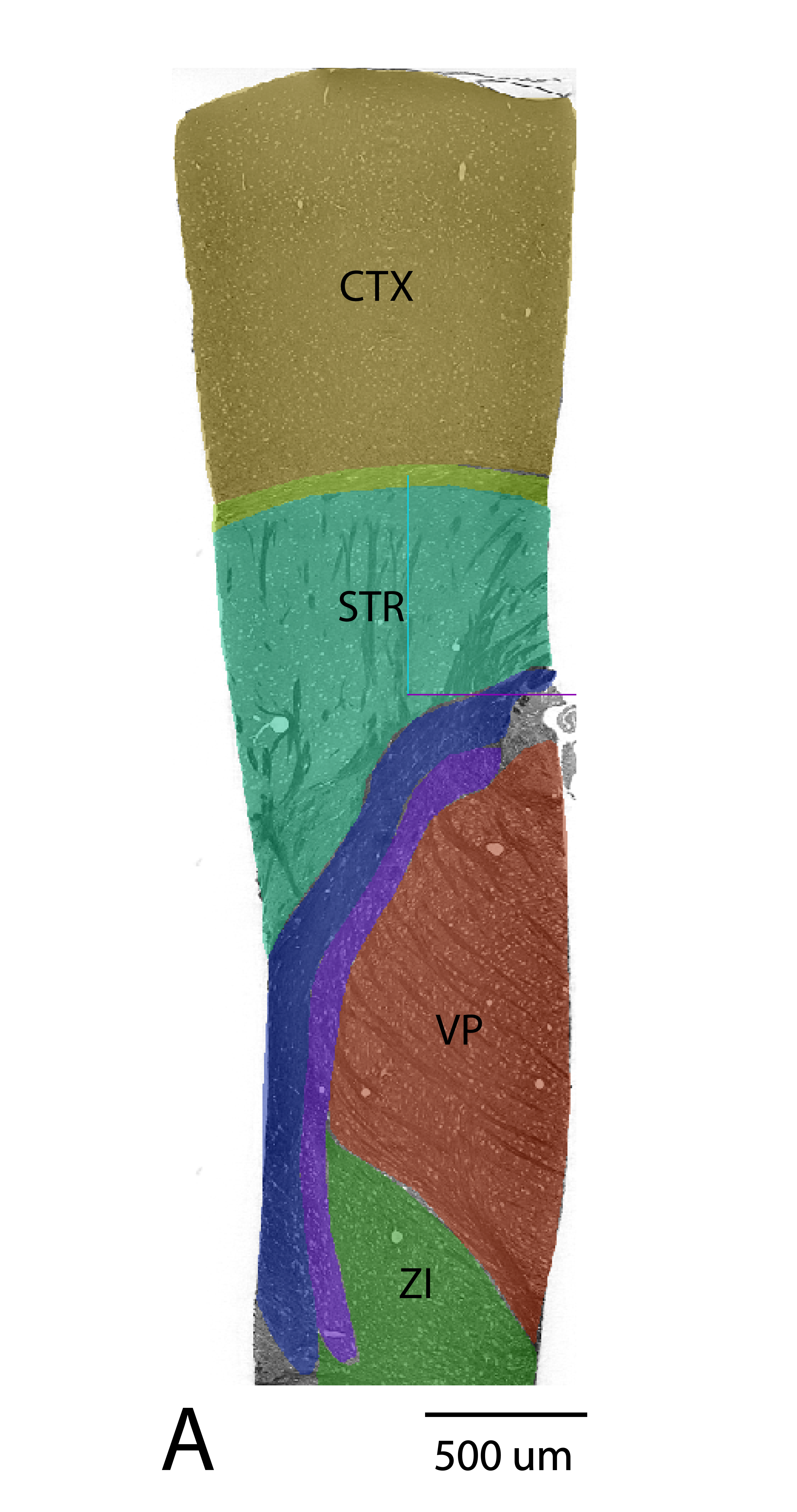}
    \end{minipage}
    %\hfill
    \begin{minipage}[b]{0.8\textwidth}
    \centering
        \includegraphics[width=\textwidth, trim={10.7cm 10.5cm 0.6cm 0.3cm},clip]{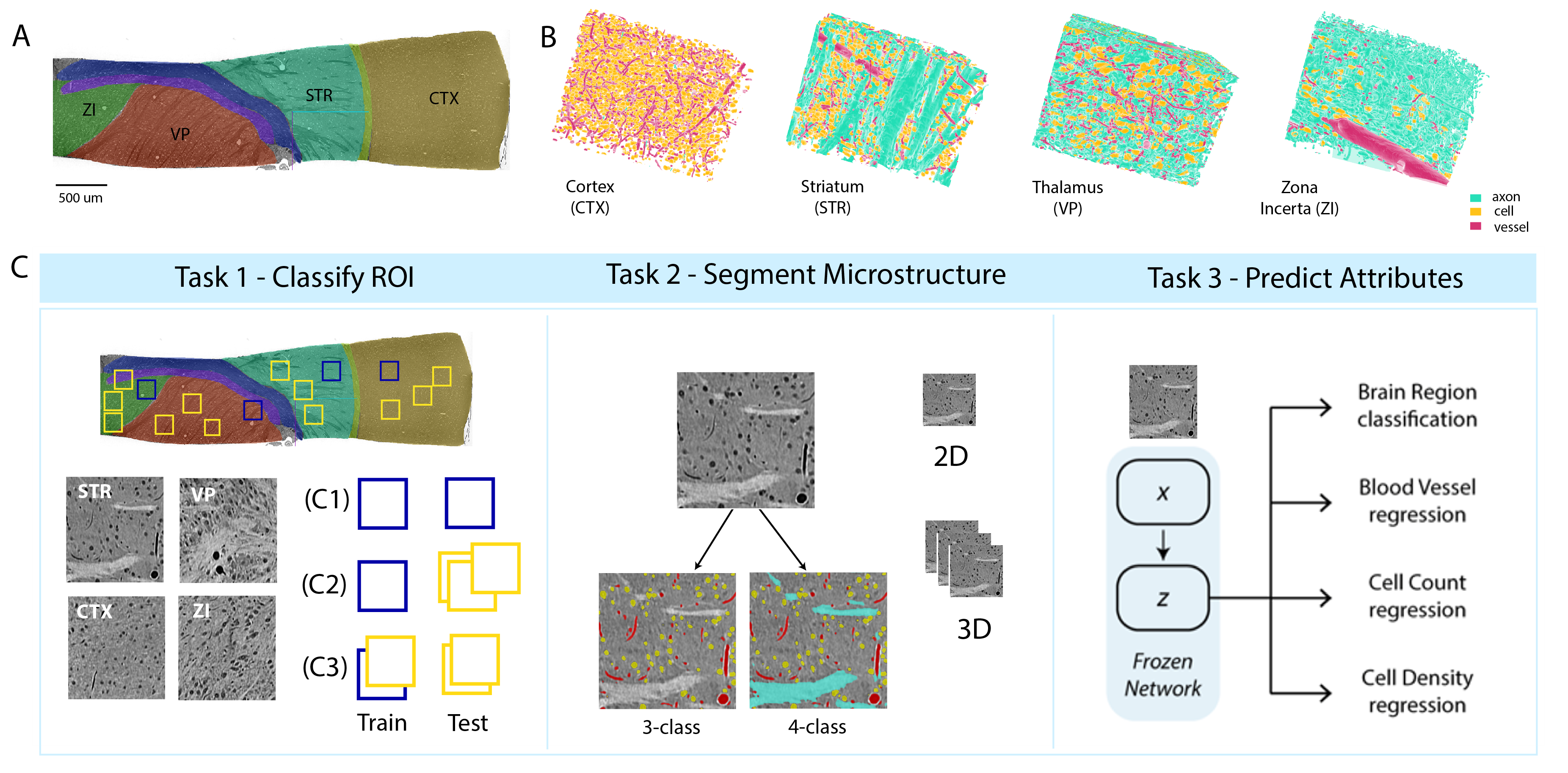}
        \raggedright
        \includegraphics[width=\textwidth, trim={0cm, 7.4cm, 0cm, 0cm,clip}]{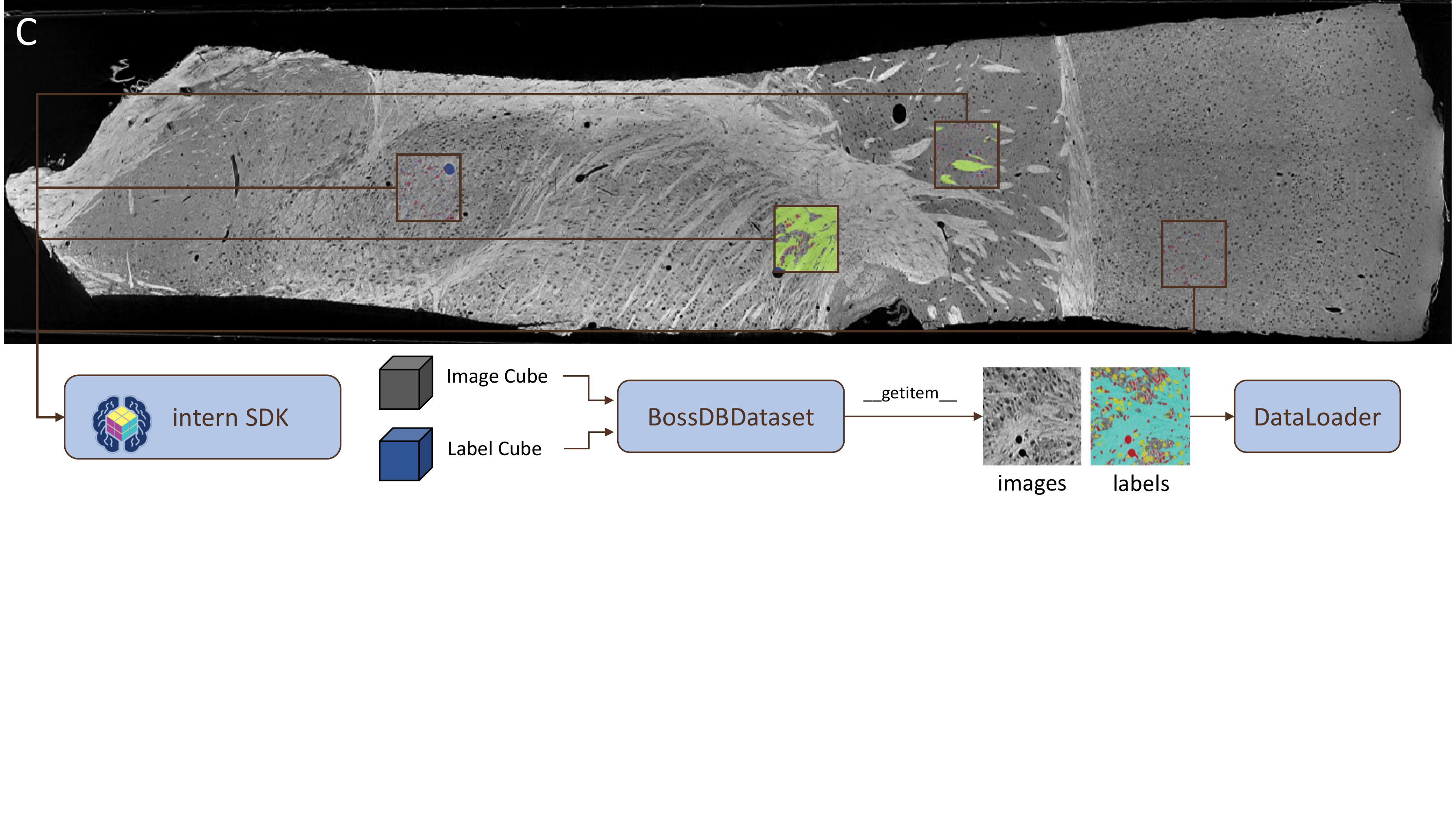}
    \end{minipage}

    \caption{\label{fig:overview} \footnotesize {\em Overview of datasets and annotations used in \ours\hspace{-1mm}. } In A we show how the dataset spans multiple brain areas including the somatosensory cortex (CTX), striatum (STR), thalamus (VP), and zona incerta (ZI). Each of these areas contains annotations of pixel-level microstructures like axons, blood vessels, and cells visualized as dense X-ray microCT volumetric imaging data (1.17 micron isotropic resolution), as seen in B.  
    The pipeline and BossDB data access are shown in C: channels are accessed through the Intern API to access cutouts without the need to download the entire dataset. This enables creating a data loader for specific task-relevant cutouts.}

\vspace{-5.5mm}
\end{figure*}

\edit{
We build our benchmark on a large open access high-resolution (1.17µm isotropic) 3D X-ray microtomography imaging dataset that provides fine-scale information about brain microstructure as well as diverse regions of interest that give more distinct global attributes \cite{prasad2020three}. The dataset thus contains a uniquely rich set of both {\em macroscale} (region of interest) and {\em microscale} (cells, blood vessels, axons) structures that can be interrogated throughout the dataset \cite{balwani2019modeling, balwani2021multi}.

The full volumetric dataset provides micron resolution of an intact brain sample totalling   $5805 \times 1420 \times 720$ pixels. The dataset spans four regions of interest: somatosensory cortex (CTX), striatum (STR), ventral posterior region of thalamus (VP), or the zona incerta (ZI) (see Figure \ref{fig:overview} A-B) and provides pixel-level microstructure (cell, axons, blood vessels, background) and macrostructure labels over 2D slices distributed over the dataset. 

To provide more data for training and validation, we expanded the pixel-level microstructure and macrostructure labels provided in the original data resource \cite{prasad2020three} to larger contiguous volumes in the data. To expand the pixel-level labels, we trained a 4-class Unet model on the sparse 2D annotations and had a trained expert proofread the annotations to create dense pixel-level microstructural labels (see Figure \ref{fig:overview}B), identifying each point as either part of an axon, cell, blood vessel, or background. The final curated pixel-level labels span 4 ROIs, with each ROI consisting 360 256x256 densely labelled images. 

To examine semantic attributes of the different images, in Task 3, we leveraged the dense pixel-level labels to compute a number of semantic features from the reconstructions: (i) the density of blood vessels, (ii) axon density, (iii) the number of cells, (iv) size of cells, and (v) the average inter-cell distance in each slice. These semantic labels provide information into different features of the cytoarchitecture that can be used to interpret the embeddings learned by models tested on this dataset. In addition to these microstructure annotations, we have expanded the macrostructure annotations from the original dataset to include interpolations of the region labels across all 720 slices of size $5805 \times 1420$. From these interpolated sections, we extracted 12 new subvolumes (three for each of the four regions of interest) for examining the generalization of models in Task 1.  }

\subsection{Data access}
The dataset and all corresponding labels are stored in BossDB \cite{hider2022brain}, the Brain Observatory Storage Service and Database. The dataset project page can be found at: \url{http://bossdb.org/project/prasad2020}. BossDB is a specialized spatial database for Electron Microscopy and X-Ray Microtomography Datasets, with seamless visualization through Neuroglancer, which enables interactive visualization of large-scale 3D annotated volumes and annotations. All data are available publicly, using public log on credentials (no account creation required). The project page documents project metadata, citation instructions, and the data creators and curators. 

For benchmarking, data are accessed through the Python intern API \cite{matelsky2020intern}. This API allows a remote connection to the BossDB system, including downloads of arbitrary, on-demand 3D cutouts of data, including raw images and annotations, without the need to download the entire dataset to disk. To facilitate the use of this API, we provide a Pytorch DataLoader for rapid algorithm development and testing. We also provide sample Jupyter notebooks to demonstrate how the task cutouts can be dowloaded, and saved as Numpy files for development in other frameworks. The tasks are defined with task-specific JSON files specifying the metadata for each task. The use of this dataloader is illustrated in Fig. \ref{fig:overview}C, where we detail how the dataset can be efficiently accessed through BossDB.

The data are structured into channels. Raw images, macrostructure annotations, and microstructure annotations, each have their own separate channel. This is detailed below. 

    {\bf Raw images}: All tasks utilize the same raw images, which are single color (grayscale) with \edit{8-bit unsigned integer} values. The total raw data volume is $720 \times 1420 \times 5805$ voxels at a resolution of $1.17\mu m$. This data is available through the raw images channel:  \url{https://api.bossdb.io/v1/mgmt/resources/prasad/prasad2020/image}.
    
     {\bf Macrostructure annotations}: Tasks 1 and 3 utilize dense pixel-level annotations of  different brain regions (macrostructure) : CTX (label 0); STR (label 1); VP (label 2); ZI (label 3). \edit{At this scale/level, there is an equal number of samples of each class and hence the classes are balanced.} This label data is represented with \edit{64-bit unsigned integers} and is accessible through the macrostructure annotations channel:
     \url{https://api.bossdb.io/v1/mgmt/resources/prasad/prasad_analysis/roi_labels}.
     
   {\bf Microstructure annotations}: Task 2 and Task 3 utilize pixel-level annotations of brain microstructure. Volumes of $256 \times 256 \times 360$ are densely annotated with microstructure labels. The labels are 0: no label (background); 1: blood vessels; 2: cells; 3: mylenated axons. Represented with \edit{64-bit unsigned integers}, it is accessible through the microstructure annotations channel: \url{https://api.bossdb.io/v1/mgmt/resources/prasad/prasad_analysis/pixel_labels}.

\edit{ A key aspect of our approach is portability of data access and training infrastructure across neuroimaging datasets in BossDB. This allows for easy extension of code and baselines to new volumetric imaging datasets stored in BossDB. These include data from new species, with new microstructure labels (synapses, membranes, mitochondria), and with new macrostructure labels (brain regions, experimental state). By modifying the dataloader JSON, developers can specify different BossDB datasets, spatial regions, and annotation (label) sources. This allows for the flexibility required to support the different tasks in this dataset, and will enable further training and deployment of these baseline models to new datasets. This is an important contribution of this work towards developing machine learning tools for emerging large-scale neuroimaging datasets. }

\subsection{Tasks}
\label{sec:tasks}

\subsubsection{Task 1: Image-level classification of brain region-of-interest (ROI)}
When analyzing imaging data that spans many different brain regions, one important question is the degree to which global image features correlate with the brain region from which the sample is drawn \cite{balwani2021multi}.
%whether or not we can determine which brain region our sample is spanning. 
Thus, we can pose this as a classification problem, where we pull a small patch (or small region) from the data and estimate which of the 4 brain regions the sample was drawn from. Given the abundance of samples available for this task across the entire volume, we can sub-divide this task into three different training schemes (see Figure \ref{fig:tasks}), detailed below. Taken together, these three training schemes allow us to evaluate how different types of models are able to generalize as more data becomes available during training.

\vspace{-3mm}
    \paragraph{ROI-C1.} For this sub-task, we use the 4 densely-annotated cubes (shown in blue in Figure \ref{fig:tasks}, Task 1), each corresponding to one of the regions of interest (CTX, STR, VP, or ZI). We divide each of the four subvolumes by selecting the first 300 ($256 \times 256$) images slices for training, and the last 50 images for testing, leaving 10-slices between the train and test data to avoid any structural overlap. \edit{ The resulting overall sample size for this sub-task is thus 1400 images, with 1200 images for the train set and 200 for the test set.}
    
    \vspace{-3mm}
    \paragraph{ROI-C2.} In this sub-task, we evaluate the performance of the models when tested on new areas within the larger 3D context of the dataset. We extract two additional $256 \times 256 \times 360$ cubes per class (shown in Figure \ref{fig:tasks}, Task 1), to serve as additional test data for the models. \edit{ This sub-task employs 4080 images, with 1200 in the train set, and 2880 in the test set.}

\vspace{-3mm}
    \paragraph{ROI-C3.} In this sub-task, we evaluate the performance of the models when allowed to learn from a larger set of data. We extracted one additional cube per class to serve as an additional source of training data for the models and
use the same test set as that employed in ROI-C2. \edit{This sub-task employs 5520 images, with 2640 in the train set, and 2880 in the test set}. 

\subsubsection{Task 2: Pixel-level segmentation of microstructures}
Another important requirement for a comprehensive mapping of brain data is correctly identifying brain microstructures such as cells and blood vessels. In this task, we utilize the dense pixel-level microstructure labels of the volumetric cutouts from 4 brain regions of interest (macrostructure; Cortex, Striatum, VP \& ZI), and evaluate different baselines on their prediction accuracy in classifying each pixel from test volumes across the brain regions into appropriate brain microstructure classes (blood vessel, cell, axon \& background). 

\vspace{-2mm}
\paragraph{3-class segmentation task.}
We first consider the pixel-level segmentation of images into one of three classes: either cell bodies, blood vessels, or other (background and axons). We use the same train and test split as in ROI-C1 in Task 1 on the main subvolumes that are densely annotated at the pixel-level (300 images for train, 50 for test, with a gap of 10 slices between datasets). This sub-task employs 1200 images for training, and 200 images for testing. 

\vspace{-2mm}
\paragraph{4-class segmentation task.}
In this task, we consider the pixel-level segmentation of images into one of four classes: either cell bodies, blood vessels, background or axons. When we consider dense axonal segmentation, we remove the ZI region from our training and testing set due to the difficulty to reliably segment axons in this subvolume even for human annotators. This sub-task employs 900 images for training, and 150 images for testing. 

\subsubsection{Task 3: Probing multiple semantic features from learned image-level embeddings}
In this task, we wanted to explore the possibility of decoding  semantic and human-interpretable features from the image-level representations learned in Task 1. If possible, it could provide a way to build interpretable image-level feature maps that contain information about microstructure without needing the expensive pixel-level labels necessary to compute these attributes in most brain mapping settings.

Specifically, we try to predict the following global properties of an image: (i) blood vessels density, (i) cell count, (iii) average cell size, (iv) axon density, and (v) average distance between cells, all through a simple linear readout. 
In the case of supervised models, these models are trained to classify images into its respective brain region; However, in the case of SSL methods, where the region-level labels aren't used to guide learning, it may be possible that other global attributes of the images may be encoded in the latent space of the model.

\section{Results}
\label{sec:results}

\subsection{Task 1: Image-level classification of brain ROI}

\begin{table}[t!]
\centering
\footnotesize
\caption{\footnotesize {\em Results on image classification accuracy for brain region prediction (Task 1).}  
%In (C1), we train and test in one subvolume with the test being held out forward slices not seen in the model. In (C2), we report the test accuracy of our models in (C1) on two new subvolumes unseen during train. In (C3), we train on a new subvolume and use the same test set in (C2). 
\footnotesize
\label{table:task1}
%We provide reach decoding accuracies for two different sequence lengths ($l=6$, $l=2$).
}
\vspace{.5mm}
\begin{tabular}{c|c|c|c}

  & \textit{ROI - C1} & \textit{ROI - C2} & \textit{ROI - C3}\\

\hline
Supervised  & 0.88 $\pm$ 0.03 & 0.77 $\pm$ 0.03 & 0.88 $\pm$ 0.02 \\
Sup w/ Mixup   & 0.90 $\pm$ 0.04 & 0.78 $\pm$ 0.03 & 0.90 $\pm$ 0.02   \\ 
%EfficientNet  & 0.97 $\pm$ & &  $\pm$ \\
\hline
BYOL    & 0.88 $\pm$ 0.02  & 0.76 $\pm$ 0.02 & 0.97 $\pm$ 0.01  \\
MYOW    & 0.90 $\pm$ 0.02 & 0.78 $\pm$ 0.05 & 0.98 $\pm$ 0.01 \\
MYOW-m   & 0.94 $\pm$ 0.02 & 0.78 $\pm$ 0.03  & 0.98 $\pm$ 0.01 \\
%BYOL (All)    & 0.86 $\pm$ 0.06 & 0.95 $\pm$ 0.02 & 0.99 $\pm$ 0.01 & 0.97 $\pm$ 0.01 \\
%MYOW (All)     & 0.85 $\pm$ 0.06 &  0.94 $\pm$ 0.02 & 0.98 $\pm$ 0.01 & 0.97 $\pm$ 0.01 \\
%MYOW-m (All)    & 0.88 $\pm$ 0.03& 0.96 $\pm$ 0.02 & 0.99 $\pm$ 0.01 & 0.98 $\pm$ 0.01\\
\hline
PCA & 0.59 & 0.25 & 0.07\\
NMF & 0.62 & 0.27 & 0.50\\

%BYOL (All)   & \multicolumn{2}{c|}{ 0.81 $\pm$ 0.03} & \multicolumn{2}{c}{} \\
%MYOW (All)   & \multicolumn{2}{c|}{ 0.80 $\pm$ 0.03} & \multicolumn{2}{c}{} \\
%MYOW-m (All) & \multicolumn{2}{c|}{0.82 $\pm$ 0.03} & \multicolumn{2}{c}{} \\
\hline
\end{tabular}
\vspace{-4mm}
\end{table}

\vspace{-2mm}
\paragraph{Experiment setup.}
%Details on architectures

In this task, we consider the classification of different images into a number of candidate brain areas (CTX, STR, VP and ZI) using representations learned through supervised and self-supervised approaches. All of the models in this task are trained using a Resnet18 encoder \cite{he2016deep}. We benchmarked two supervised  models: the first trained using standard approaches for regularization (20\% dropout and weight decay factor of 0.3) and the other trained using mixup \cite{zhang2017mixup}. Given the underlying shared structure of image samples in volumetric data, we also consider a number of  self-supervised learning methods suitable for this task: (i) BYOL \cite{grill2020bootstrap}, (ii) MYOW \cite{azabou2021mine}, (iii) and a variant of MYOW that we tested with a single projector and predictor (MYOW-merged, or MYOW-m). For the SSL models, we follow the standard procedure of freezing the network weights after training, and then training a linear layer on top of the representations.  This tells us how well the SSL loss captures the classes in the data after only a linear transformation. All models have a latent dimensions of $256$. As additional baselines, we also extracted 256-dimensional embeddings from our data using Principal Component Analysis (PCA) and Non-Negative Matrix Factorization (NMF), and trained a linear layer on these representations. 

In our experiments, we evaluate all methods across 5 training instances with different random seeds, and report the overall mean accuracy and standard deviation. All models are trained for 100 epochs using an SGD optimizer with a learning rate of 0.03. For more details on our experimental setup and models, see Section~\ref{app:task1} in the Appendix.

\vspace{-2mm}
\paragraph{Results in classifying brain ROIs.}
\label{sec:res-small-scale}

We report the results of the three subtasks for ROI classification in Table \ref{table:task1}. In our first subtask (ROI-C1), we find that many of the SL and SSL models achieve comparable accuracy, with the MYOW-m model achieving the highest accuracy in this limited training regime. %At the same time, we find that our simple unsupervised baselines (PCA, NMF) do not provide separable classes.
In ROI-C2, we test the generalization capabilities of the models trained in ROI-C1 by evaluating how well they performed on subvolumes in other parts of the larger dataset (Table~\ref{table:task1}, ROI-C2). There is considerable heterogenity across brain regions and thus we can consider this as a form of domain generalization.
%In this case, we test on three held out volumes in each ROI (12 subvolumes in total in test). 
In this case, we observe a significant decrease in accuracy that can be attributed to the domain shift in data. 
%\paragraph{ROI-C3.}
In ROI-C3, we add another set of training data (see Figure~\ref{fig:overview}) and test on the same volumes as in our last subtask. In this case, we can observe that SSL methods significantly outperform the SL models, with SSL models achieving even higher accuracy than in the small-scale case (97-98\%) and supervised models achieving a much more modest improvement (88-90\%) with the new training data.
We can thus observe a significant  gap in SSL over SL models, showing that exposure to additional data (in this case, the additional cube with respect to task ROI-C1) drastically improves the generalization of the SSL models to unobserved data.

\subsection{Task 2: Pixel-level segmentation of microstructures}

In Task 2, we consider four different variants of pixel-level segmentation. The first variant employs 2D models for pixel-level segmentation. The second variant employs 3D models instead. The third variant is a 4-class setting where ZI is removed from the brain regions involved (as axons in this region are difficult to distinguish accurately for even human annotators). The fourth variant is a 3-class setting where all 4 brain regions (Cortex, Striatum, VP and ZI) are utilized but only 3 classes are considered (blood vessels, cells, background+axons; avoids axon segmentation). 

\vspace{-2.5mm}
\paragraph{Experiment setup.}

In our experiments, we perform pixel-level segmentation using a selected set of 2D and 3D models. Each model is put through a separate hyper-parameter tuning process for finding an optimal learning rate and batch size \edit{ by training on the train set and evaluating on the validation split}. Each model is trained for 20 epochs with its optimal learning rate and batch size and evaluated across 5 training instances (each with its own random seed). The class-wise F1-score, the class-wise IoU, and the overall mean and standard deviation of both metrics are reported for each model (Table~\ref{tab:task2}). We do not report accuracy because for this particular task of pixel-level segmentation, it does not aptly represent the model performance \edit{due to class imbalance (see Appendix Section \ref{app:class-prop} for breakdown of different classes in the training and test sets)}. 

The models we used for the 2D segmentation task are the standard 2D U-Net model \cite{ronneberger2015unet,schmidt2019stdunet} and selected models from the 'segmentation\_models.pytorch' library \cite{Yakubovskiy:2019}: MAnet \cite{fan2020manet}, FPN \cite{lin2017fpn}, U-Net++ \cite{zongwei2018unetpp}, PAN \cite{li2018pan} and PSPNet \cite{zhao2016pspnet}. The models we used for the 3D segmentation task are the standard 3D U-Net model \cite{ronneberger2015unet} and selected models from `MedicalZooPytorch' \cite{adaloglou2019MRIseg}: VNetLight \cite{milletari2016vnet, adaloglou2019MRIseg} and HighResNet \cite{li2017highresnet}. For more details refer Section~\ref{app:task2} in the Appendix.

%%%%%%%%%% TABLE %%%%%%%%%%

\begin{table}[t!]
\centering
\footnotesize
\caption{{\em F1 \& IoU scores for models trained on the pixel-level segmentation task (Task 2).\label{table:seg}} \\
%\footnotesize
}\addtolength{\tabcolsep}{-3pt}
\resizebox{\columnwidth}{!}{

\begin{tabular}{cc|cccc|ccccc}
\multicolumn{11}{c}{\em I.  2D Pixel-level Segmentation}\\
  \multicolumn{2}{c}{\textit{}} & \multicolumn{4}{c}{\textit{3-Class}} & \multicolumn{5}{c}{\textit{4-Class}}\\
Method & Metric & Bg + Axons & Vessels & Cells & Avg. & Bg & Vessels & Cells & Axons & Avg.\\ 
\hline
2D U-Net & F1 & 0.99 & 0.76 & 0.85 & $0.87 \pm 0.012$ & 0.97 & 0.82 & 0.87 & 0.94 & $0.90 \pm 0.003$\\
2D U-Net & IoU & 0.98 & 0.64 & 0.75 & $0.79 \pm 0.014$ & 0.89 & 0.70 & 0.77 & 0.60 & $0.74 \pm 0.008$ \\
\hline
MA-Net & F1   & 0.99 & 0.79 & 0.87 & $0.88 \pm 0.003$ & 0.97 & 0.83 & 0.87 & 0.94 & $0.90 \pm 0.002$ \\
MA-Net & IoU   & 0.98 & 0.68 & 0.78 & $0.81 \pm 0.003$ & 0.89 & 0.71 & 0.78 & 0.76 & $0.78 \pm 0.011$ \\
%\hline
%Linknet (F1)  & 0.99 & 0.78 & 0.87 & $0.88 \pm %0.002$ & 0.96 & 0.82 & 0.86 & 0.94 & $0.90 \pm %0.004$ \\
%Linknet (IoU)   & 0.98 & 0.67 & 0.77 & $0.81 \pm %0.003$ & 0.88 & 0.70 & 0.76 & 0.78 & $0.78 \pm %0.006$ \\
\hline
FPN & F1   & 0.99 & 0.72 & 0.84 & $0.85 \pm 0.01$ & 0.96 & 0.73 & 0.84 & 0.93 & $0.86 \pm 0.004$\\
FPN & IoU   & 0.97 & 0.59 & 0.73 & $0.76 \pm 0.015$ & 0.87 & 0.59 & 0.72 & 0.72 & $0.72 \pm 0.021$ \\
\hline
U-Net++ & F1   & 0.99 & 0.79 & 0.87 & $0.89 \pm 0.002$ & 0.97 & 0.81 & 0.85 & 0.93 & $0.89 \pm 0.015$ \\
U-Net++ & IoU   & 0.98 & 0.68 & 0.78 & $0.81 \pm 0.002$ & 0.88 & 0.68 & 0.75 & 0.73 & $0.76 \pm 0.036$ \\
\hline
PAN & F1    & 0.98 & 0.60 & 0.80 & $0.79 \pm 0.035$ & 0.95 & 0.69 & 0.80 & 0.93 & $0.84 \pm 0.007$ \\
%0.95 & 0.68 & 0.80 & 0.93 & $0.84 \pm 0.005$ & 
PAN & IoU   & 0.96 & 0.46 & 0.66 & $0.69 \pm 0.039$ & 0.85 & 0.53 & 0.67 & 0.76 & $0.70 \pm 0.014$ \\
%0.84 & 0.52 & 0.66 & 0.77 & $0.70 \pm 0.010$ & 
\hline
PSPNet & F1   & 0.97 & 0.48 & 0.74 & $0.73 \pm 0.013$ & 0.94 & 0.54 & 0.71 & 0.91 & $0.78 \pm 0.012$ \\
PSPNet & IoU   & 0.94 & 0.39 & 0.61 & $0.65 \pm 0.043$ & 0.82 & 0.38 & 0.55 & 0.74 & $0.62 \pm 0.015$ \\
\hline \hline
\multicolumn{11}{c}{\em{II.  3D Pixel-level Segmentation}}\\
   \multicolumn{2}{c}{\textit{}} & \multicolumn{4}{c}{\textit{3-Class}} & \multicolumn{5}{c}{\textit{4-Class}}\\
Method & Metric & Bg + Axons & Vessels & Cells & Avg. & Bg & Vessels & Cells & Axons & Avg.\\ 
\hline
3D U-Net & F1 & 0.99 & 0.77 & 0.87 & $0.88 \pm 0.006$ & 0.93 & 0.76 & 0.80 & 0.87 & $0.84 \pm 0.032$ \\
%0.89 & 0.76  & 0.81 & 0.85 & $0.83 \pm 0.049$ \\
3D U-Net & IoU & 0.98 & 0.65 & 0.76 & $0.80 \pm 0.007$ & 0.81 & 0.62 & 0.67 & 0.50 & $0.65 \pm 0.045$ \\
%0.76 & 0.61 & 0.68 & 0.53 & $0.65 \pm 0.045$ \\
\hline
VNetLight & F1   & 0.99 & 0.75 & 0.83 & $0.85 \pm 0.012$ & 0.90 & 0.65 & 0.73 & 0.76 & $0.76 \pm 0.063$ \\
%0.67 & 0.30  & 0.63 & 0.73 & $0.58 \pm 0.125$ \\
%& 0.99 & 0.74 & 0.83 & $0.85 \pm 0.011$ & 0.67 & 0.30  & 0.63 & 0.73 & $0.58 \pm 0.125$ \\
VNetLight & IoU  & 0.97 & 0.61 & 0.70 & $0.76 \pm 0.013$ & 0.78 & 0.46 & 0.58 & 0.43 & $0.56 \pm 0.061$ \\
%0.70 & 0.36  & 0.57 & 0.46 & $0.42 \pm 0.103$ \\
%& 0.97 & 0.67 & 0.70 & $0.76 \pm 0.011$ & 0.70 & 0.36  & 0.57 & 0.46 & $0.42 \pm 0.103$ \\
\hline
HighResNet & F1   & 0.99 & 0.74 & 0.84 & $0.85 \pm 0.019$ & 0.89 & 0.51 & 0.73  & 0.77 & $0.72 \pm 0.083$ \\
%& 0.99 & 0.74 & 0.85 & $0.86 \pm 0.013$ & 0.88 & 0.51 & 0.72  & 0.79 & $0.72 \pm 0.067$ \\
HighResNet & IoU   & 0.97 & 0.61 & 0.72 & $0.77 \pm 0.026$ & 0.73 & 0.35 & 0.58  & 0.42 & $0.52 \pm 0.075$ \\
%& 0.97 & 0.61 & 0.73 & $0.77 \pm 0.015$ & 0.70 & 0.36 & 0.57  & 0.46 & $0.52 \pm 0.051$ \\

%\hline
%#Model 4 (F1)   & X & X & X & X & X & X  & X & X & X \\
%Model 4 (IoU)   & X & X & X & X & X & X  & X & X & X \\
\hline
\end{tabular}
}
\label{tab:task2}
\addtolength{\tabcolsep}{3pt}
\vspace{-1mm}
\end{table}

%%%%%%%%%% FIGURE %%%%%%%%%%
\begin{figure*}[ht!]
\centering
\includegraphics[width=0.99\textwidth,trim={1mm 0cm 0cm 0cm},clip]{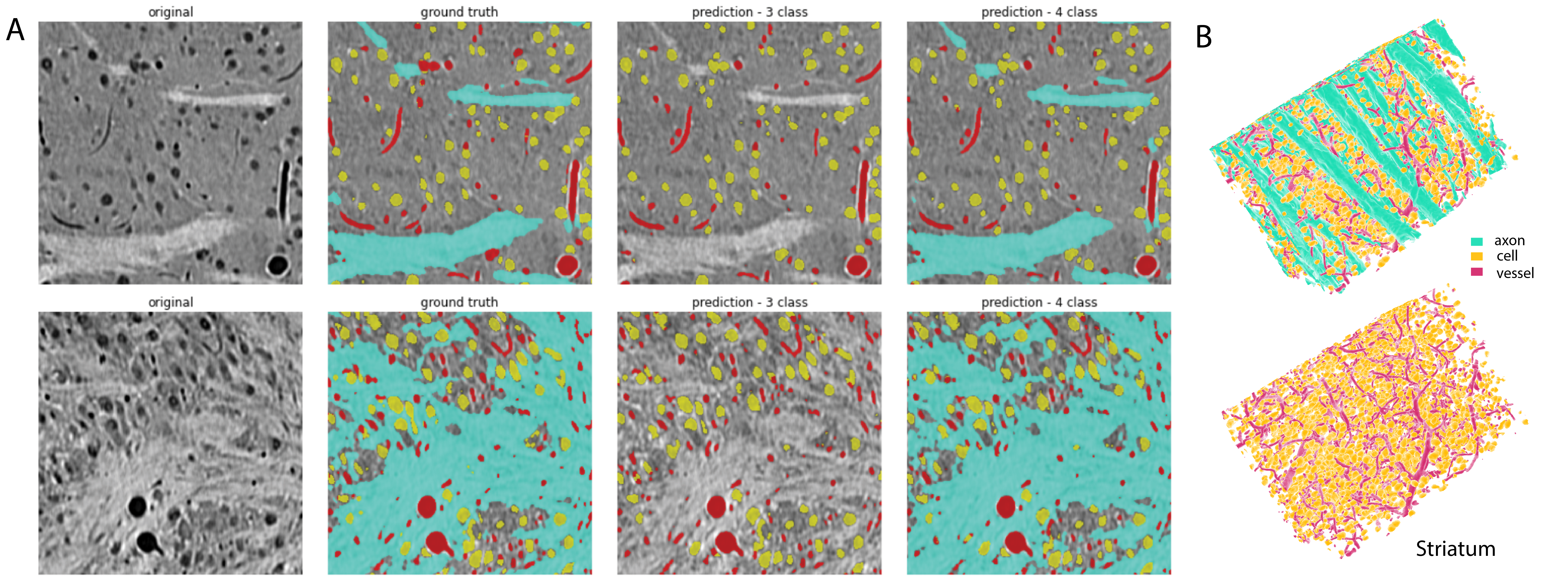}
\caption{\footnotesize {\em Visualization of predictions from pixel-level segmentation task (Task 2).} From left to right: original image, ground truth overlaid, prediction from U-Net model in the 3-class  and 4-class settings (top row is Striatum, bottom is VP). Axons are visualized in cyan, blood vessels in red, and cell bodies are in yellow. B) 3D reconstruction from the predictions for Striatum for the 4-class (top) and 3-class (bottom) settings.\vspace{-3mm}}
\label{fig:seg}
\end{figure*}

%\begin{table}[]
%\centering
%\begin{tabular}{c|c}
%                                  & accuracy         \\ \hline
%SegNet           & 0.69         \\ \hline
%U-Net              & 0.75  \\ \hline
%{\color{red} Other baselines?}                 & \\ \hline
%
%\end{tabular}
%\caption{\footnotesize{{\em Results from Task 2: Pixel-level segmentation accuracy (2D) } {\color{red} From MTL paper} \vspace{-2mm}}}
%\label{tab:synaps}
%\end{table}

%\begin{table}[]
%\centering
%\begin{tabular}{c|c}
%                                  & accuracy         \\ \hline
%SegNet           & x        \\ \hline
%U-Net              & x  \\ \hline
%{\color{red} Other baselines?}              %   & \\ \hline

%\end{tabular}
%\caption{\footnotesize{{\em Results from %Task 2: Pixel-level segmentation accuracy %(3D) } {\color{red} From MTL paper} %\vspace{-2mm}}}
%\label{tab:synaps}
%\end{table}

\vspace{-2.5mm}
\paragraph{Pixel-level segmentation in 2D.}
The results from the selected models on our 2D pixel-level segmentation task are tabulated in Part I of Table~\ref{tab:task2} and visualized in Figure~\ref{fig:seg}. The individual slices are the input to the models and they are fed in a batched manner during training. For training and evaluation, we consider both the 3-class and 4-class settings and we use EfficientNet-b7 encoder for all the models as it was seen to give the best performance among the 25 encoders that were attempted. From our results we see that MA-Net performed the best overall among the 2D models with an average IoU of 0.78 followed by U-Net++ with an average IoU of 0.76 in the 4-class setting (Table~\ref{tab:task2}). As can be seen from the class-wise IoU and class-wise F1-scores, for most of the best performing models, the most challenging components to differentiate are cells and blood vessels, which are also difficult for human annotators to identify from 2D slices without further 3D context. Also, we can note, the average IoU (across classes and models) increases from 0.72 to 0.79 upon moving from 4-class to 3-class setting. This indicates an expected performance improvement as there are more slices (more data) and fewer classes (easier) in the 3-class setting.

\vspace{-2mm}
\paragraph{Pixel-level segmentation in 3D.} 
We provide the same breakdown and accuracy measures as in the 2D case, for this 3D case (Part II of Table~\ref{tab:task2}). For providing 3D input to the models, we pass in consecutive slices (8 slices) as a single subvolume and build a prediction over all slices jointly in 3D. The U-Net model performed best with 0.80 IoU (3-class) which is only competent with the best 2D models. This would be because we define the subvolumes in a non-overlapping manner, so even though we have improved 3D context, the model also sees fewer samples/inputs. Due to the same reason, moving from 3-class to 4-class setting, we can note a drop in performance since there are fewer samples. Larger sample counts / allowing overlap could enable 3D models to perform better.

\subsection{Task 3: Probing multiple semantic features from learned image-level embeddings}

\vspace{-2mm}
In Task 3, we explore the prediction of different semantic attributes (e.g., density of blood vessels or cells) from the latent space learned by the models trained in Task 1. %This task provides an important test to understand whether different attributes of the microarchitecture are embedded in the representations 

\begin{table}[b]
\centering
\footnotesize
\caption{\footnotesize \label{tab:task3} {\em Task 3: $R^2$ scores on multi-task feature readout for supervised and SSL models trained in Task 1.} \vspace{2mm}
\footnotesize
\label{table:decoding}
%We provide reach decoding accuracies for two different sequence lengths ($l=6$, $l=2$).
}
\vspace{.5mm}
\begin{tabular}{c|c|c|c|c|c}

\multicolumn{6}{c}{\em I. Linear Readouts from Models Trained on a Single Subvolume (ROI-C1)}\\
  Methods & Vessels & Axons & Cell Count & Cell Size  & Dist (k=1) \\
  \hline
{Supervised}   & 0.77 $\pm$  0.06    &  0.94 $\pm$ 0.01  & 0.67 $\pm$ 0.06  & 0.61 $\pm$ 0.05 & 0.48 $\pm$ 0.05 \\
{Sup w/ Mixup}  & 0.82 $\pm$ 0.02 & 0.95 $\pm$ 0.00 & 0.71 $\pm$ 0.02 & 0.67 $\pm$ 0.03 & 0.47 $\pm$ 0.02  \\
\hline
{BYOL}   &  0.85 $\pm$ 0.01  & 0.94 $\pm$ 0.01 &  0.75 $\pm$ 0. 01 &  0.69 $\pm$  0.01 & 0.49 $\pm$ 0.01  \\
{MYOW}  & 0.85 $\pm$ 0.01 & 0.94 $\pm$ 0.01 & 0.74 $\pm$  0.01  & 0.69 $\pm$  0.01  & 0.50 $\pm$ 0.02  \\
{MYOW-m}  & 0.87 $\pm$ 0.01 & 0.95 $\pm$ 0.01 & 0.77 $\pm$  0.01  & 0.69 $\pm$  0.01  & 0.51 $\pm$ 0.01  \\
\hline
{ PCA}   & 0.75  &  0.82 & 0.55 & 0.47 &  0.31   \\
{ NMF}   &  0.81 & 0.85 &  0.59 & 0.55 &  0.34  \\
\hline

\multicolumn{6}{c}{\em II. Linear Readouts from Models Trained on Two Subvolumes (ROI-C3)}\\
  Methods & Vessels & Axons & Cell Count & Cell Size  & Dist (k=1) \\
  \hline
{Supervised}   & 0.79 $\pm$ 0.02  & 0.94 $\pm$ 0.02 & 0.73 $\pm$ 0.02 & 0.63 $\pm$ 0.04 & 0.49 $\pm$ 0.02  \\
{Sup w/ Mixup}  &  0.75 $\pm$ 0.04 & 0.88 $\pm$ 0.04  & 0.64 $\pm$ 0.04  & 0.54 $\pm$ 0.07 & 0.37 $\pm$ 0.05   \\
\hline
{BYOL}   & 0.88 $\pm$ 0.00  & 0.96 $\pm$ 0.00  & 0.79 $\pm$ 0.00 &  0.73 $\pm$  0.01 & 0.53 $\pm$ 0.02    \\
{MYOW}  & 0.88 $\pm$ 0.01 & 0.96 $\pm$ 0.00  & 0.79 $\pm$ 0.01  & 0.72 $\pm$ 0.01 & 0.52 $\pm$ 0.01  \\
{MYOW-m}  & 0.87 $\pm$ 0.01  & 0.96 $\pm$ 0.01 & 0.78 $\pm$ 0.01 & 0.72 $\pm$ 0.01  & 0.53 $\pm$ 0.01   \\
\hline
{ PCA}   & 0.75  & 0.82 &  0.53 &  0.46 & 0.29  \\
{ NMF}   & 0.75   & 0.83  & 0.56 &  0.49 &  0.31  \\
\hline

% \multicolumn{11}{c}{} \\ % empty line to give some space

%\multicolumn{5}{c}{\textit{II. 3D Semantic Readout}}  \vspace{0.5mm} \\ 

%{\color{blue} SSL models}   &  & & & \\
%\hline
\end{tabular}
%\vspace{-1mm}
\end{table}

\begin{figure*}[ht!]
\centering
\vspace{-1mm}
\includegraphics[width=0.98\textwidth ,trim={0cm 22.0cm 0cm 0cm},clip]%
{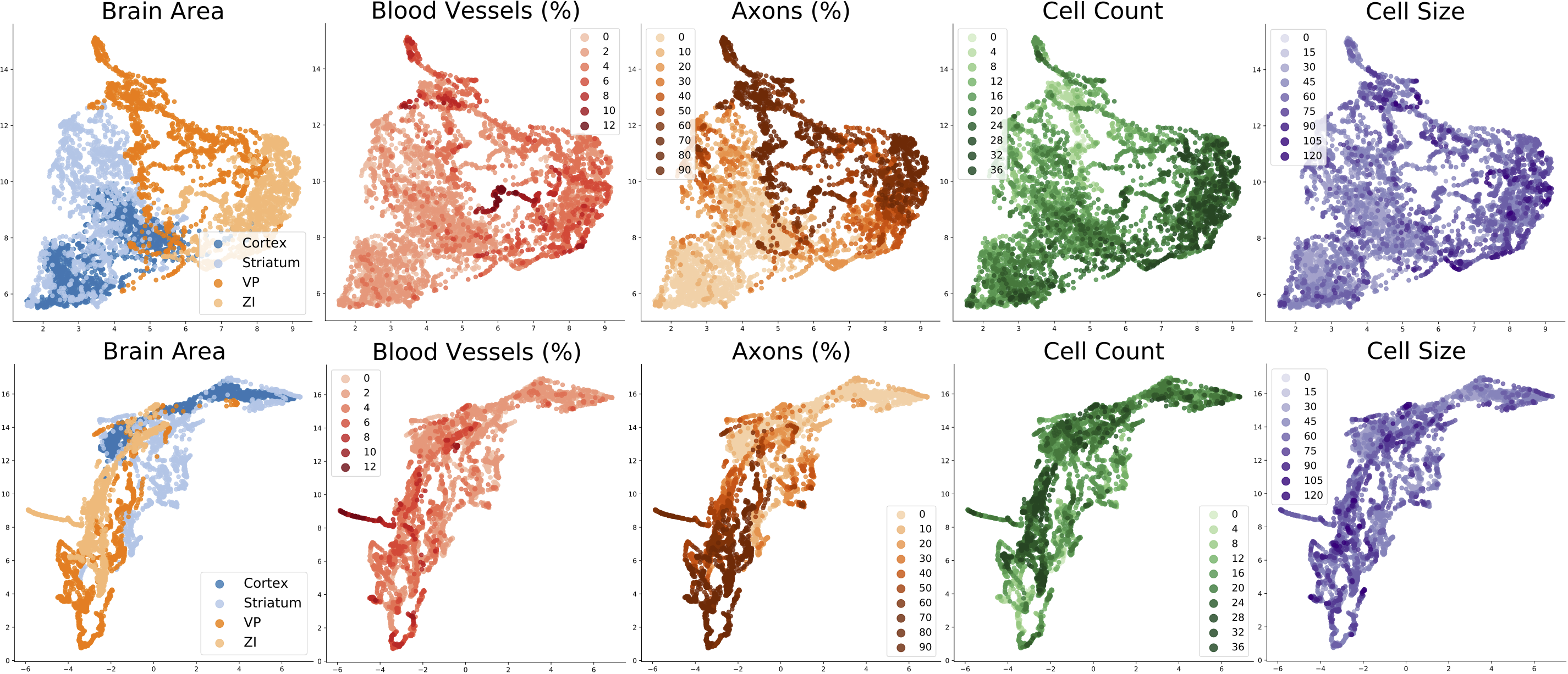}
\caption{\footnotesize {\em Visualization of the representations learned by MYOW-m in Task 1 (ROI-C1) with global semantic attributes for each image visualized as different colors.} From left to right, latents are colored by brain area (class), \% blood vessels, \% axons, cell count, and cell size.
%The top row shows   embeddings obtained by training  on a single subvolume task (ROI-C1) while the bottom row shows  embeddings obtained after training on two subvolumes (ROI-C3). 
%The first column represents these embeddings labelled with different brain regions. Other columns overlay these embeddings with the density of blood vessels,  density of axons, cell counts, and cell sizes. It can be seen that different embedding regions are associated with different semantic features such as brain regions, axon density etc.
\vspace{-2mm}}
\label{fig:comb_latent}
\end{figure*}

To do this, we leveraged the high-quality dense microstructure annotations to extract information about the attributes of each image that included: the proportion of pixels that are either blood vessels or axons,  the cell count and size, and average distance between cells and their nearest neighbor. Further details on the experimental setup for this task can be found in  the Appendix~in Section~\ref{app:task3}.

%understanding how relevant these are for a given task, and can open up studies on how neuroanatomy informs models when trained without a specific downstream task in mind (SSL methods).
We use the trained models from Task 1, freeze their weights, and compute the representations of all of the annotated images used in ROI-C1 (both train and test, 360 slices). From these latent features, we fit a simple linear regression model on these representations to predict each of the  desired semantic attributes. We have limited data for this task and thus report the $R^2$ on all of the latents (no train and test split in this case). 

We  report the $R^2$ values in Table \ref{tab:task3} for all 360 slices in the densely annotated volume used in Task 1 and 2 for the models trained in ROI-C1 (Table~\ref{tab:task3} I) and on more data in ROI-C3 (Table~\ref{tab:task3} II). In both cases, we find that the SSL models significantly outperform the rest of the baseline models. The gap is more pronounced in the two volume training condition, highlighting the generalization difference between supervised and SSL approaches \cite{TENDLE2021Generalizability} in providing good representations for a wide range of tasks.  In Figure \ref{fig:comb_latent}, we project the MYOW-m (ROI-C1) embeddings using UMAP \cite{mcinnes2018umap}, and overlay the semantic features.

\vspace{-2mm}
\section{Discussion}
\label{sec:conclusion}
In this paper, we introduced a multi-task benchmark for analysis of brain structure from high-resolution neuroimaging data spanning many brain areas. \edit{ In addition, we also built general infrastructure for training models from datasets in the BossDB framework, like dataloaders which can be adapted to different datasets. This will expand the use of large-scale volumetric neuroimaging data for machine learning tool development.}

{\bf Limitations and future work.} In neuroscience, often obtaining high quality labeled data (especially for dense segmentation tasks like those provided in Task 2), is very costly \cite{Rolnick2019Generative}. The intensive nature of manual annotation and proofreading data thus limits the amount of labeled and annotated data that we provide to train models on, or the amount of distributional shift that can be assessed. While this is a limitation of the work, it also is an accurate reflection of the challenges faced in the field, and thus requires more label-efficient approaches for learning like the SSL methods we highlight. Moving forward, we hope to leverage the models tested in this work to generate even more high quality annotated data to further improve model performance and segmentation.

When designing our current benchmark, we focused on building a multi-scale challenge where variability was due to changes in brain structure and not different preparations or imaging parameters. Therefore we focused on a single animal where we have a large intact brain volume that spans many heterogeneous brain regions. While this may limit the generalization of the models to new samples or datasets, it also addresses the heterogeneous nature of different brain regions that is often overlooked. Our results show that even within a single brain, there is rich heterogeneity across different brain regions that makes it difficult for some models to generalize. This also reveals important generalization gaps between SL and SSL models.

In the future, we hope to expand this effort to learn models of brain structure from other non-convolutional architectures (e.g., point cloud-based models of neural structure \cite{jackson2022building}), and deploy our tools on new multi-scale brain datasets, perhaps using new lightsheet  \cite{Hillman2019Light} or whole-brain scaling \cite{Trinkle2021Synchrotron, Foxley2021Multi} techniques.  %Also, we anticipate hosting a challenge on this benchmark with community participation.

{\bf Broader impacts.}
The high heterogeneity of brain data, coupled with the variability in the scale and nature of the tasks presented in this benchmark make it a challenging and useful resource for the broader ML community. Furthermore, we hope the provided dataset and tasks, as well as the supporting codebase and BossDB infrastructure will help accelerate development of ML techniques for emerging, high-resolution neuroimaging datasets being collected in the broader community.

\section*{Acknowledgements}
This project was supported by  NSF award  IIS-2039741, NIH award 1R01EB029852-01, NIH award R01MH126684, NIH award R24MH114785 in addition to generous gifts from the Alfred Sloan Foundation, the McKnight Foundation, and the CIFAR Azrieli Global Scholars Program.

\bibliographystyle{IEEEbib}

\bibliography{main}

\newpage
\setcounter{section}{0}
\setcounter{table}{0}
\renewcommand{\thetable}{A\arabic{table}}

\setcounter{figure}{0}
\renewcommand{\thefigure}{A\arabic{figure}}

\section*{\Large Appendix}

%%%%%%%% DATASET  %%%%%%%
\section{Dataset Details}
\label{app:data}
\subsection{Spatial autocorrelation between images}

\begin{figure*}[h!]
\centering
\includegraphics[width=0.9\textwidth,trim={0 0 0 0cm},clip]{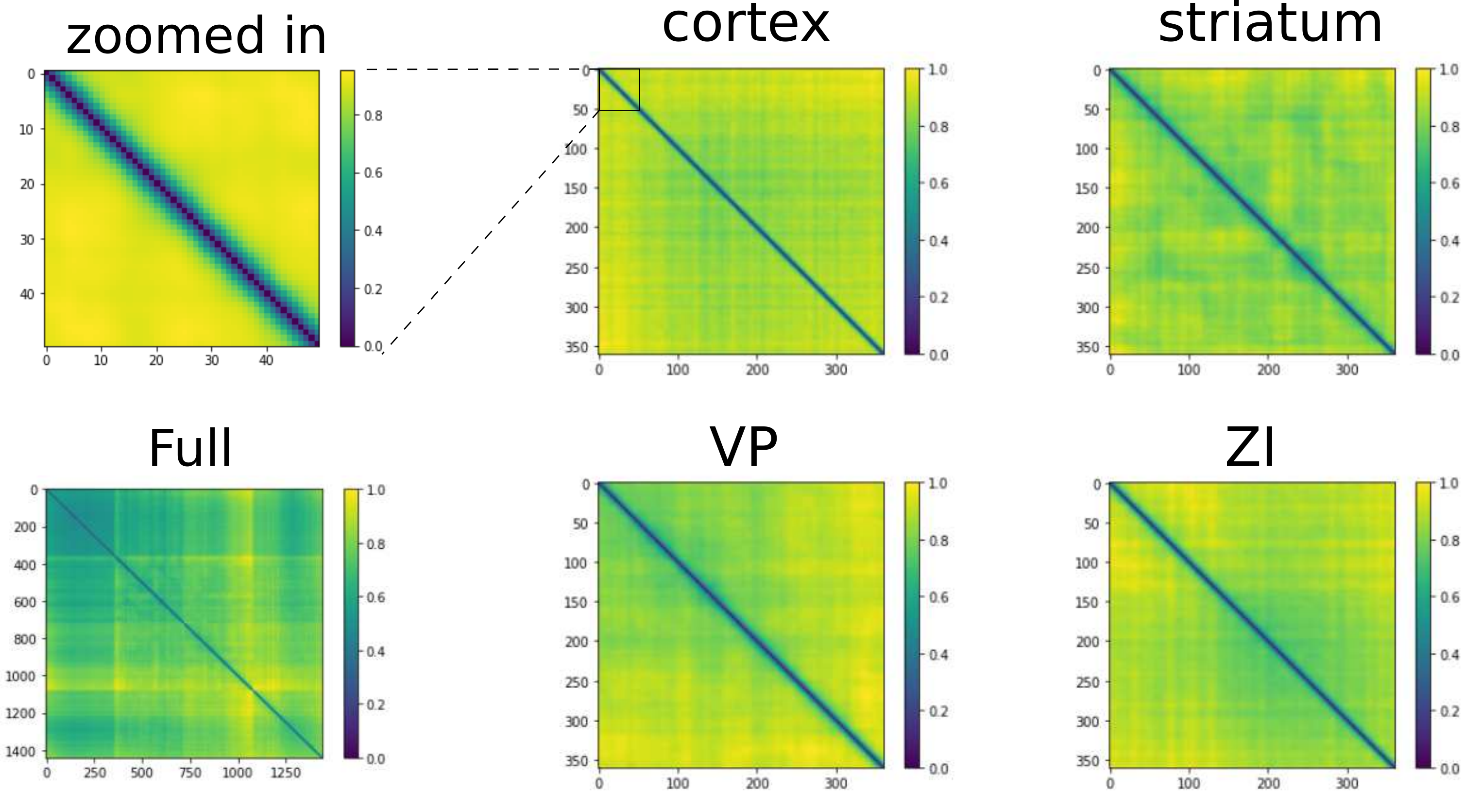}
\caption{\footnotesize {\em Cross-covariance matrices.} Per-class cross-covariance of all 4 considered regions (right), zoomed-in view of cortex cross-covariance(top left) and cross-covariance of entire dataset (bottom left).
\vspace{-3mm}}
\label{fig:crosscorr}
\end{figure*}

We computed the cross covariance of the images in our main subvolumes to see how much correlation exists between consecutive or nearby image slices. This analysis confirmed that after 5-7 slices, the cross-correlation between images drops off consistently (see Figure \ref{fig:crosscorr}). Based upon this analysis, we crop out 10 slices between the training and testing sets in all the subvolumes.

\subsection{Dense pixel-level annotations}

Utilizing the sparse annotations from \cite{prasad2020three}, we trained a 2D UNet to segment the image data into 4 classes (blood vessels, cell bodies, axons, and background) and applied it to the slices in the four volumes which were not already annotated. A proofreader then reviewed the annotations for each slice in the volume using ITK-Snap \cite{py06nimg} in the y-z plane. During the proofreading process of a slice, any structure with labels that were split between classes (errors in the UNet) were first corrected. Then, any structures without split labels were reviewed for correctness and changed if incorrect. Finally, the annotator checked the following Z slice to ensure continuity of components. This process continued for each slice in the volume. To ensure consistency of the annotations, the volume was proofread by the same annotator a second time, now going through the x-z plane. After proofreading we generated densely annotated volumes of size 256x256x360 in each of the four ROIs.

\subsection{Interpolated ROI annotations}

A full ROI annotation of the x-ray microtomography dataset was interpolated from manually annotated z-level slices that were roughly 50 pixels apart. To generate the interpolated layers, a recursive algorithm was implemented to linearly shift the boundary between two annotation classes from one pre-existing layer to the next: once a shifting boundary region was located, the midpoint of the shift was identified and full, one voxel-wide annotations were created along the midpoint spanning all necessary z levels. This process split transition regions into two roughly equal spaces, in which the recursive process repeated for the two respective transition spaces until the entire transition region was fully annotated.

However, this process was not able to account for all border transitions. Thus, about 10\% of the border transitions were annotated manually using the visualization software application ITK-SNAP. Fully interpolated ROI annotations are available for z-levels between 109 and 459, inclusive.

\begin{figure*}[h!]
\centering
\includegraphics[width=0.9\textwidth]{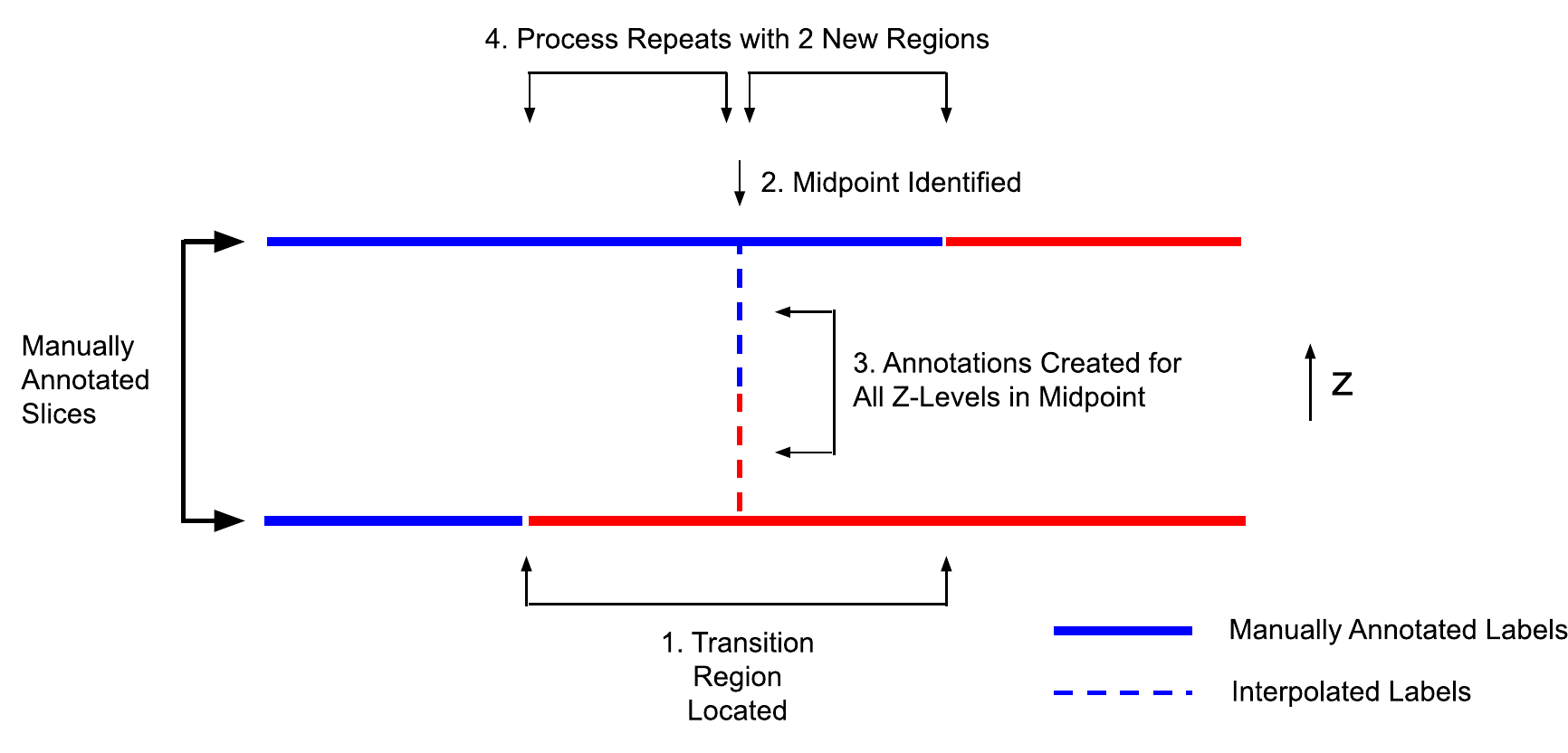}
\caption{\footnotesize {\em The Recursive Process of the Interpolation of ROI Annotations.} A four step recursive process was implemented to produce interpolated ROI annotation layers from manually annotated layers. This process recursively repeated until the entire transition region was interpolated with ROI annotations. 
\vspace{-3mm}}
\label{fig:latent}
\end{figure*}

\section{Data Access}

\label{app:boss}
To allow for easy, publicly accessible data, the dataset is stored in the BossDB system. Critically, this database enables efficient access of arbitrary cutouts of large volumetric datasets. This project page is available at \url{https://bossdb.org/project/prasad2020}. No user account is required. Public credentials will allow read-only access of the data without the need for user account creation. 

The raw images are available with the resource identifier \url{bossdb://prasad/prasad2020/image}, which is the ``prasad'' experiment, ``prasad2020'' experiment, and ``image'' channel. The raw image data are in an unsigned 8-bit integer format, with preferred indexing in $ZYX$ format. Areas with invalid data (e.g. outside the image volume) are assigned the value 0. The maximum and minimum indices are $z=[0,720]$, $y=[0,1420]$, and $x=[0,5805]$. The data have $1.17\mu m$ isotropic resolution.  

The annotation images are available with the resource identifier \url{bossdb://prasad/prasad_analysis/pixel_labels} and \url{bossdb://prasad/prasad_analysis/roi_labels}, which is the ``prasad'' collection, ``prasad\_analysis'' experiment, and ``pixel\_labels'' and ``roi\_labels'' channels. The annotation data are in an unsigned 64-bit integer format, with preferred indexing in $ZYX$ format. Areas with invalid data (e.g. outside the image volume) are assigned the value 0. As for the raw images channel, the maximum and minimum indices are $z=[0,720]$, $y=[0,1420]$, and $x=[0,5805]$. The data have $1.17\mu m$ isotropic resolution. 

Access is available through the Python intern array API, which provides numpy-like referencing. Using this library, the user creates an intern array, \verb|image_array = array(boss_url)|, where \verb|boss_url| is one of the resource identifiers listed above. The user can then access data from the channel using the numpy-like syntax, demonstrated by this code example to download a cutout corresponding to cortex.  
\begin{verbatim}
from intern import array
image = array(boss_url)
data = image[110:379, 900:1156, 4600:4856]
\end{verbatim}

A data loader is provided for rapid development of Pytorch models, which is used for all models tested in this work. On creation, the data loader loads a task configuration `.json' file which specifies the parameters of the task. Two modes of operation are allowed, \verb|download=true| and \verb|download=false|. For the former, on creation of the data loader, data are pulled locally and stored as a numpy file. If the numpy file already exists, it is instead loaded from disk. In the later case, data are downloaded to memory and not written to disk. The data are stored in a numpy tensor containing the concatenated slices from cortex, striatum, vp, zi (for four class problems). The item retrieval function can serve up integer region labels (for task 1), or microstructure masks (for task 2), as dictated by the configuration. The basic input data configuration parameters are:
\begin{itemize}
\item \textbf{image\_chan} - BossDB channel from which to pull the raw image data.
\item \textbf{annotation\_chan} - BossDB channel from which to pull the image annotations (either macro- or micro-structures).
    \item \textbf{xrange\_cor} - range along the x-axis (on the full data) for the slices from Cortex.
    \item \textbf{yrange\_cor} - range along the y-axis (on the full data) for the slices from Cortex.
    \item \textbf{xrange\_stri} - range along the x-axis (on the full data) for the slices from Striatum. 
    \item \textbf{yrange\_stri} - range along the y-axis (on the full data) for the slices from Striatum. 
    \item \textbf{xrange\_vp} - range along the x-axis (on the full data) for the slices from VP.
    \item \textbf{yrange\_vp} - range along the y-axis (on the full data) for the slices from VP. 
    \item \textbf{xrange\_zi} - range along the x-axis (on the full data) for the slices from ZI. 
    \item \textbf{yrange\_zi} - range along the y-axis (on the full data) for the slices from ZI. 
    
    \item \textbf{z\_train} - the range (slices) along the z-axis to use as the train set.
    \item \textbf{z\_val} - the range (slices) along the z-axis to use as the val set.
    \item \textbf{z\_test} - the range (slices) along the z-axis to use as the test set. A buffer of 10 slices is recommended between the train/val set and the test set.
    \item \textbf{volume\_z} - the number of slices to include in a volume slice. For the 2D cases, the individual slices are the input to the models (number of slices in a volume slice is 1). In the 3D case, volume slices are the input to the models.
\end{itemize}

The repository, \url{https://github.com/MTNeuro/MTNeuro}, contains scripts and python notebooks for data access and running models. Python notebooks which download the numpy files for images and annotations are provided for users not using Pytorch. These are based on the original data access notebooks developed for this dataset \url{https://github.com/nerdslab/xray-thc}. The repository and requirements can be installed via the \verb|pip| package manager. The repository is structured as
\begin{itemize}
    \item MTNeuro: the core code for running pytorch dataloaders and pytorch models
    \item Notebooks: access notebooks for users who do not use pytorch
    \item Scripts: example scripts for running pytorch models for each task, which form the basis for new algorithm development 
\end{itemize}

%%%%%%%%%%  TASK 1  %%%%%%%%%% 
\section{Further Details on Task 1}
\label{app:task1}

\subsection{Models}
\subsubsection{Supervised models}

\begin{itemize}
    \item Resnet18: 18-layer version of a model that reformulates the layers as learning residual functions with reference to the layer inputs in order to efficiently train deeper architectures \cite{he2016deep}.
    \item Resnet18 + Mixup: same model architecture described above (Resnet18), but trained using mixup, an augmentation that mixes inputs as well as their corresponding labels through a convex combination, and has been shown to yield significant improvement in supervised classification models \cite{zhang2017mixup}
\end{itemize}

\subsubsection{Self-supervised models}

We test several contrastive-based SSL methods that do not need negative examples to generate good representations of data \cite{grill2020bootstrap}. All SSL models are trained using a Resnet18 encoder backbone in order to make them comparable to the evaluated supervised models.

\begin{figure*}[h!]
\centering
\includegraphics[width=0.7\textwidth]%
{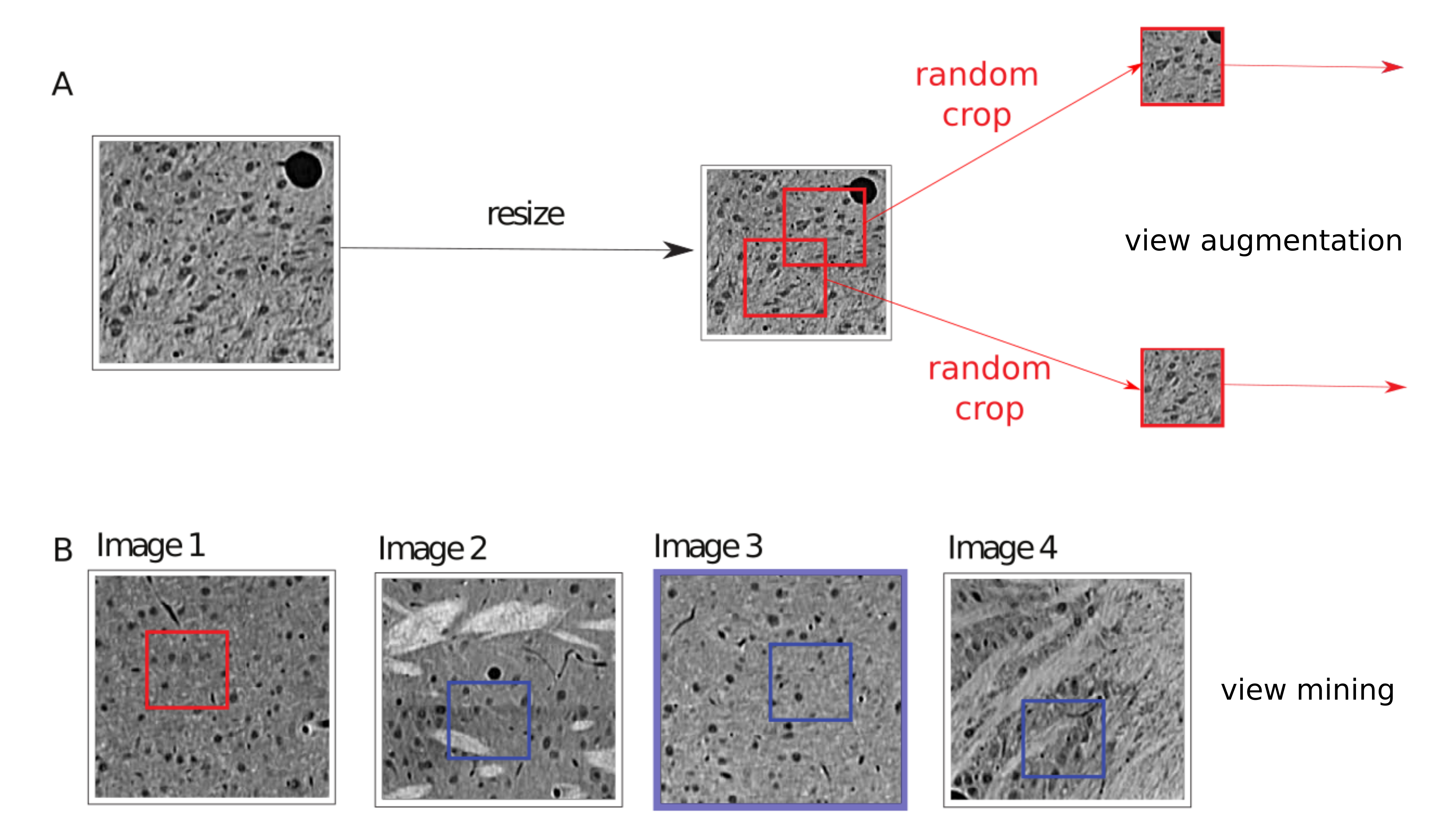}
\caption{\footnotesize {\em View generation for self-supervised approaches.}  In A we show how augmented views are generated by extracting random crops from a given image sample. In B, we schematize the view mining process, involving comparing different random crops from a pool of samples to a target view (red) and choosing the closest one in representation space as the mined view (Image 3, highlighted).}
\label{fig:augs}
\end{figure*}

We tested the following SSL models:
\begin{itemize}
    \item BYOL: BYOL relies on a mirrored structure of online and target networks that learn from each other \cite{grill2020bootstrap}. In particular, the online network tries to predict the encoding of a target image view $x$ by minimizing the distance in latent space to an augmented view $x_a$.
    \item MYOW: MYOW builds on BYOL by also minizing the distance in latent space between views, but incorporates a view mining approach in order to search the dataset for neighboring samples in representations space. 
    The augmented and mined samples are then integrated into a unified latent space through the use of an additional predictor network \cite{azabou2021mine}.
    \item MYOW-m: MYOW-m is an extension of MYOW that integrates MYOW's sample mining procedure, but uses a single predictor on both the augmented and mined views rather than using the cascaded projector design proposed in MYOW.
\end{itemize}
Since all of these SSL methods rely on some form of transformation to process the views during training, we choose the simplest possible augmentation for all three methods: randomly cropping patches half the width and height of each image sample to generate augmented and mined views (see Figure \ref{fig:augs}). In order to standardize the evaluation of the methods, we evaluate the trained encoder's performance in classifying which class each of the four corners of an image sample belongs to.

\subsection{More Details on Training}
\subsubsection{Configuration Files}
Two types configuration files are provided as input for training:
\begin{itemize}
    \item \textbf{Network Configuration} - a '.json' file containing settings for the optimizer, augmentations to use, the parameters for the corresponding supervised or self-supervised models, and the seed to use for the training.
    \item \textbf{Task Configuration} - a '.json' file specifying information regarding the data slices to access for the training. This includes the database channel to access for the data and  region-level annotations, the x and y ranges for the slices from each region, the size of the slices (corresponding to the level of downsampling applied), the train-val-test split, the depth of each slice sample (1, since we focus on 2D processing for this task). 
\end{itemize}

\subsubsection{Model Settings}
The basic model settings are:
\begin{itemize}
    \item \textbf{model} - either the classifier (if supervised) or encoder backbone (if SSL) to train for the classification task.
    \item \textbf{classes} - the number of output classes of the model.
    \item \textbf{method} - (only for SSL) the self-supervised approach to train.
\end{itemize}

\subsubsection{Input Data Configuration}

The dataset configurations used for the results in Table~\ref{table:task1}, column ROI-C1 are listed in Table~\ref{tab:data_config}. The coordinates for the additional cubes used in ROI-C2 and ROI-C3 are shown in  Table~\ref{tab:data_extra_cubes}.
\begin{table}
    \centering
    \begin{tabular}{|c|c|}
        \hline
        \textbf{Setting} & \textbf{Values cube set A} \\
        \hline
        xrange\_cor & [4600,4856] \\
        yrange\_cor & [900,1156]  \\
        xrange\_stri & [3700,3956] \\
        yrange\_stri & [500,756] \\
        xrange\_vp & [3063,3319] \\
        yrange\_vp & [850,1106] \\
        xrange\_zi & [1543,1799] \\
        yrange\_zi & [650,906] \\
        z\_train & [110, 379] \\
        z\_val & [380, 409] \\
        z\_test & [420, 470] \\
        \hline    
    \end{tabular}
    \vspace{3pt}
    \caption{\footnotesize {\em Dataset configuration used for task 1 (ROI-C1) and task 2}}
    \label{tab:data_config}
\end{table}

\begin{table}
    \centering
    \begin{tabular}{|c|c|c|c|}
        \hline
        \textbf{Setting} & \textbf{Values cube set B}& \textbf{Values cube set C}& \textbf{Values cube set D} \\
        \hline
        xrange\_cor & [5312,5568] & [5056,5312] & [4600,4856]\\
        yrange\_cor & [388,644] & [644,900] & [400,656] \\
        xrange\_stri & [3828,4084]  & [3344,3600] & [3800,4056] \\
        yrange\_stri & [912,1168] & [400,656] & [244,500] \\
        xrange\_vp & [2551,2807]  & [2151,2407] & [2295,2551]  \\
        yrange\_vp & [850,1106] & [950,1206] & [694,950] \\
        xrange\_zi & [1287,1543] & [1031,1287] & [1799,2055] \\
        yrange\_zi & [906,1162] & [906:1162] & [650:906]  \\
        \hline    
    \end{tabular}
    \vspace{3pt}
    \caption{\footnotesize {\em Coordinates for the additional cube sets used in ROI-C2 (C and D) and ROI-C3 (B, C and D)}}
    \label{tab:data_extra_cubes}
\end{table}

\subsubsection{Training and evaluation setup}
We train all supervised and self-supervised models using a 0.03 learning rate, a batch size of 256 (chosen for a stable linear layer evaluation) and 5 different random seed values: 1, 100, 350, 631 and 872. We compile these 5 results in order to calculate the mean and standard deviation of the classification accuracy for each considered approach. For the Resnet18-Mixup setting, we evaluate 5 different probabilities of applying the augmentation during training: 0.01, 0.1, 0.3, 0.5 and 0.7; and report the setting with the best performing mean accuracy.

\subsection{Additional experiments}

\begin{table}[t!]
\centering
\footnotesize
\caption{\footnotesize {\em Results on image classification for brain area prediction (Task 1).} 
\footnotesize
\label{tab:task1-app}
%We provide reach decoding accuracies for two different sequence lengths ($l=6$, $l=2$).
}
\vspace{.5mm}
\begin{tabular}{c|c|c|c}
  & \multicolumn{3}{c}{\textit{ROI-C1}} \\
  & \textit{Downsampling =4} & \textit{Downsampling =2} & \textit{Full-res} \\
\hline
Resnet18 & 0.83 $\pm$ 0.06  & 0.88 $\pm$ 0.03 &  0.87 $\pm$ 0.1\\
Resnet18-Mixup   & 0.85 $\pm$ 0.06 & 0.90 $\pm$ 0.04 & 0.91 $\pm$ 0.03 \\ 
% Resnet50 & 0.83 $\pm$ 0.06  & 0.88 $\pm$ 0.03 &  0.87 $\pm$ 0.1& 0.96 $\pm$ 0.01\\
% Resnet50 & 0.97 $\pm$ 0.01  &   0.99$\pm$0.004   &  0.99 $\pm$ 0.0008\\
% EfficientNet & 0.91 $\pm$ 0.03  & 0.99 $\pm$ 0.005 & 0.78  $\pm$ 0.17 &  $\pm$ \\
%ViT & 0.71 $\pm$ 0.03  &  $\pm$  & 0.89  $\pm$ 0.01 \\
\hline
BYOL   & 0.83 $\pm$ 0.04   & 0.88 $\pm$ 0.02  & 0.98 $\pm$ 0.01  \\
MYOW    & 0.84 $\pm$ 0.04 & 0.90 $\pm$ 0.02 & 0.96 $\pm$ 0.03  \\
MYOW-m  & 0.84 $\pm$ 0.05 & 0.94 $\pm$ 0.02 & 0.99 $\pm$ 0.01   \\
\hline
PCA & 0.59 & 0.59& 0.59 \\
NMF & 0.55 & 0.62& 0.54\\
%BYOL (All)    & 0.86 $\pm$ 0.06 & 0.95 $\pm$ 0.02 & 0.99 $\pm$ 0.01 & 0.97 $\pm$ 0.01 \\
%MYOW (All)     & 0.85 $\pm$ 0.06 &  0.94 $\pm$ 0.02 & 0.98 $\pm$ 0.01 & 0.97 $\pm$ 0.01 \\
%MYOW-m (All)    & 0.88 $\pm$ 0.03& 0.96 $\pm$ 0.02 & 0.99 $\pm$ 0.01 & 0.98 $\pm$ 0.01\\
\hline %\hline
%\multicolumn{5}{c}{\textit{II. Large-scale evaluation}} \vspace{0.5mm} \\ 
%  & \multicolumn{2}{c|}{\textit{2D classification}} & \multicolumn{2}{c}{\textit{3D classification}}\\
%  \hline
%Resnet18 & \multicolumn{2}{c|}{0.96$\pm$0.0009} & \multicolumn{2}{c}{0.91$\pm$0.01}\\
%Resnet18-Mixup   & \multicolumn{2}{c|}{} & \multicolumn{2}{c}{} \\ 
%Resnet50 & \multicolumn{2}{c|}{0.98$\pm$0.002} & \multicolumn{2}{c}{ $\pm$ }\\
%EfficientNet & \multicolumn{2}{c|}{0.95$\pm$0.004} & \multicolumn{2}{c}{ $\pm$ }\\ 
%ViT & \multicolumn{2}{c|}{0.74$\pm$0.02} & \multicolumn{2}{c}{ $\pm$ }\\ \hline \hline
%BYOL    & \multicolumn{2}{c|}{0.75 $\pm$ 0.03} & \multicolumn{2}{c}{0.82 $\pm$ 0.02} \\
%MYOW     & \multicolumn{2}{c|}{0.77 $\pm$ 0.06} & \multicolumn{2}{c}{0.85 $\pm$ 0.03} \\
%MYOW-m   & \multicolumn{2}{c|}{0.76 $\pm$ 0.04} & \multicolumn{2}{c}{0.84 $\pm$ 0.02} \\
%\hline
%BYOL (All)   & \multicolumn{2}{c|}{ 0.81 $\pm$ 0.03} & \multicolumn{2}{c}{} \\
%MYOW (All)   & \multicolumn{2}{c|}{ 0.80 $\pm$ 0.03} & \multicolumn{2}{c}{} \\
%MYOW-m (All) & \multicolumn{2}{c|}{0.82 $\pm$ 0.03} & \multicolumn{2}{c}{} \\
\end{tabular}
\vspace{-1mm}
\end{table}

We report additional tests in Table \ref{tab:task1-app}, where we evaluate all considered models under the ROI-C1 training setting, but using different downsampling factors on the image samples: 4x, 2x and full resolution (no downsampling). We observe that performance increases with the resolution (as more information is available to the models). Surprisingly, we see a significantly larger performance jump in the SSL methods when moving from 2x downsampling to full resolution tests with respect to the supervised methods (which increase their performance only by a slight margin). This further supports our observation that SSL models benefit more from exposure to additional data than their supervised counterparts. Each SSL training instance took on average 30 minutes to train under 4x downsampling, 1.5 hours under 2x downsampling, and 4 hours under full resolution setting (runtime for a single random seed, trained on an RTX 3090 GPU node). We chose to use 2x downsampling for the rest of our experiments in task 1, since it provided a good compromise between performance and runtime.

%%%%%%%%%%  TASK 2  %%%%%%%%%% 
\section{Further Details on Task 2}
\label{app:task2}

\subsection{Baselines}

\subsubsection{2D Models}
The models we use for the 2D segmentation task are: 
\begin{itemize}
    \item A standard \textbf{2D U-Net} model \cite{ronneberger2015unet,schmidt2019stdunet} 
    \item Selected models from the 'segmentation\_models.pytorch' library \cite{Yakubovskiy:2019}: 
    \begin{itemize}
        \item \textbf{MAnet} - a model utilizing a multi-scale attention mechanism, originally design for liver and liver tumor segmentation \cite{fan2020manet},
        \item \textbf{FPN} - the Feature Pyramid Network architecture for object detection modified for image segmentation \cite{lin2017fpn},
        \item \textbf{U-Net++} - a nested U-Net architecture developed for medical image segmentation \cite{zongwei2018unetpp},
        \item \textbf{PAN} (Pyramid Attention Network) - a model incorporating spatial pyramid attention structure, designed to utilize global contextual information in semantic segmentation \cite{li2018pan},
        \item \textbf{PSPNet} (Pyramid Scene Parsing Network) - a model utilizing pyramid pooling and a scene parsing network to learn global context information better \cite{zhao2016pspnet}.
    \end{itemize}
\end{itemize}

\subsubsection{3D Models}
The models we use for the 3D segmentation task are:
\begin{itemize}
    \item A standard \textbf{3D U-Net} model \cite{ronneberger2015unet},
    \item Selected models from `MedicalZooPytorch' \cite{adaloglou2019MRIseg}:
    \begin{itemize}
        \item \textbf{VNetLight }- a lighter version of the V-Net convolutional network architecture developed to perform volumetric segmentation \cite{milletari2016vnet, adaloglou2019MRIseg},
        \item \textbf{HighResNet} - a compact, high-resolution convolutional network for volumetric segmentation, originally demonstrated on brain parcellation pretext task on brain MR images \cite{li2017highresnet}. 
    \end{itemize}
\end{itemize} 

\subsection{More Details on Training}
\subsubsection{Configuration Files}
Two types configuration files are provided as input to training:
\begin{itemize}
    \item \textbf{Network Configuration} - a '.json' file containing settings for the optimizer, augmentation setting, the model settings, evaluation settings, settings for saving the output and the seed to use for the training.
    \item \textbf{Task Configuration} - a '.json' file specifying information regarding the data slices to access for the training. This includes the database channel to access for the data and annotations, the x and y ranges for the slices from each region, the size of the slices, the train-val-test split, the size of the volume slice (3D) and the whether the training is for 3-Class or 4-Class setting. The task configuration used for the results in Table~\ref{tab:task2} are listed in Table~\ref{tab:data_config}.
\end{itemize}

Together, these two files completely specify the configurations for the training run.

\subsubsection{Model Settings}
The basic model settings are:
\begin{itemize}
    \item \textbf{encoder\_name} - the encoder to use with the model. This is applicable only for the 2D models and the 'UNet\_3D' model. For more information visit 'segmentation\_models.pytorch` library (\cite{Yakubovskiy:2019}). For the training results in Table~\ref{tab:task2}, 'efficientnet-b7' was used as the encoder as it gave the best performance among several other encoders that were tried.
    \item \textbf{encoder\_weights} - the pre-trained weights to use for the model. For the 2D model and 'UNet\_3D' results in Table~\ref{tab:task2}, weights trained on ImageNet were used.
    \item \textbf{in\_channels} - the number of input channels.
    \item \textbf{classes} - the number of output classes.
\end{itemize}

\subsubsection{Details on  Class Proportions}
\label{app:class-prop}
 \edit{At the microstructure level/scale, the frequency of each class across the brain regions are as follows:
    \vspace{-12pt}
    \begin{itemize}
    \itemsep0em
        \item Cortex - Background: 93\%; Blood Vessel: 2.33\%; Cell: 4.64\%; Axons: ~0\% 
        \item Striatum - Background: 72.63\%; Blood Vessel: 2.5\%; Cell: 5.66\%; Axons: 19.22\%
        \item VP -  Background: 22.75\%; Blood Vessel: 4.73\%; Cell: 5.73\%; Axons: 66.8\%
        \item ZI - Background: 32.15\%; Blood Vessel: 5.4\%; Cell: 9.41\%; Axons: 53.04\%
        \item Total - Background: 55.13\%; Blood Vessel: 3.74\%; Cell: 6.36\%; Axons: 34.77\% 
    \end{itemize} 
}

\subsubsection{Hyper-Parameters and Random Seeds}
For the selection of optimal hyper parameters for each model a hyper parameter search is performed by training the models for the following learning rates: 0.1, 0.05, 0.01, 0.005, 0.001; and the following batch sizes: 2,4,6,8,10. The best performing learning rate and batch size is chosen and 5 separate instances of each model are trained with this optimal learning rate and batch size, with seeds: 1, 100, 350, 631 and 872 respectively. These 5 results (for each model) are used to calculate the mean and SD ($Mean \pm SD$) of the performance metrics (which is reported in Table \ref{tab:task2}). Also across several models it was seen that Adam optimizer was yielding a better result than SGD, so Adam optimizer was used for all the training runs. The optimal hyperparameters found for each model in each setting are listed in Table \ref{tab:optimal_hyperparameters}.

\begin{table}[ht!]
\centering
\footnotesize
\caption{{\em Optimal Hyperparameters used for Task 2 models in Table \ref{tab:task2}} \\
%\footnotesize
}\addtolength{\tabcolsep}{-3pt}
\resizebox{0.8\columnwidth}{!}{

\begin{tabular}{c|cc|cc}
\multicolumn{5}{c}{\em I.  2D Pixel-level Segmentation}\\
  \multicolumn{1}{c}{\textit{}} & \multicolumn{2}{c}{\textit{3-Class}} & \multicolumn{2}{c}{\textit{4-Class without ZI}}\\
Model & Learning Rate & Batch Size & Learning Rate & Batch Size\\ 
\hline
2D U-Net & 0.1 & 8 & 0.01 & 4 \\
\hline
MA-Net & 0.01 & 2 & 0.001 & 8 \\
\hline
FPN & 0.01 & 2 & 0.01 & 4 \\
\hline
U-Net++ & 0.001 & 8 & 0.01 & 8 \\
\hline
PAN & 0.01 & 10 & 0.001 & 8 \\
\hline
PSPNet & 0.01 & 4 & 0.1 & 2 \\
\hline \hline
\multicolumn{5}{c}{\em{II.  3D Pixel-level Segmentation}}\\
   \multicolumn{1}{c}{\textit{}} & \multicolumn{2}{c}{\textit{3-Class}} & \multicolumn{2}{c}{\textit{4-Class without ZI}}\\
Model & Learning Rate & Batch Size & Learning Rate & Batch Size\\ 
\hline
3D U-Net & 0.005 & 1 & 0.01 & 1 \\
\hline
VNetLight & 0.01 & 1 & 0.01  & 1 \\
\hline
HighResNet & 0.005 & 1 & 0.001 & 1 \\

\hline
\end{tabular}
}
\label{tab:optimal_hyperparameters}
\addtolength{\tabcolsep}{3pt}
\vspace{-1mm}
\end{table}

%%%%%%%%%%  TASK 3  %%%%%%%%%% 
\section{Further Details on Task 3}

In order to extract the semantic features from the microstructure annotations in the 4 densely connected cubes, we first isolate the relevant corresponding pixel-level labels (either cells, blood vessels or axons). Once we have extracted the desired class and labeled everything else as background, we can perform a connected component analysis to compute the desired semantic features (cell count, size, and distance, as well as axon and vessel density) for each image sample.

\begin{figure*}[h!]
\centering
\includegraphics[width=0.98\textwidth]%
{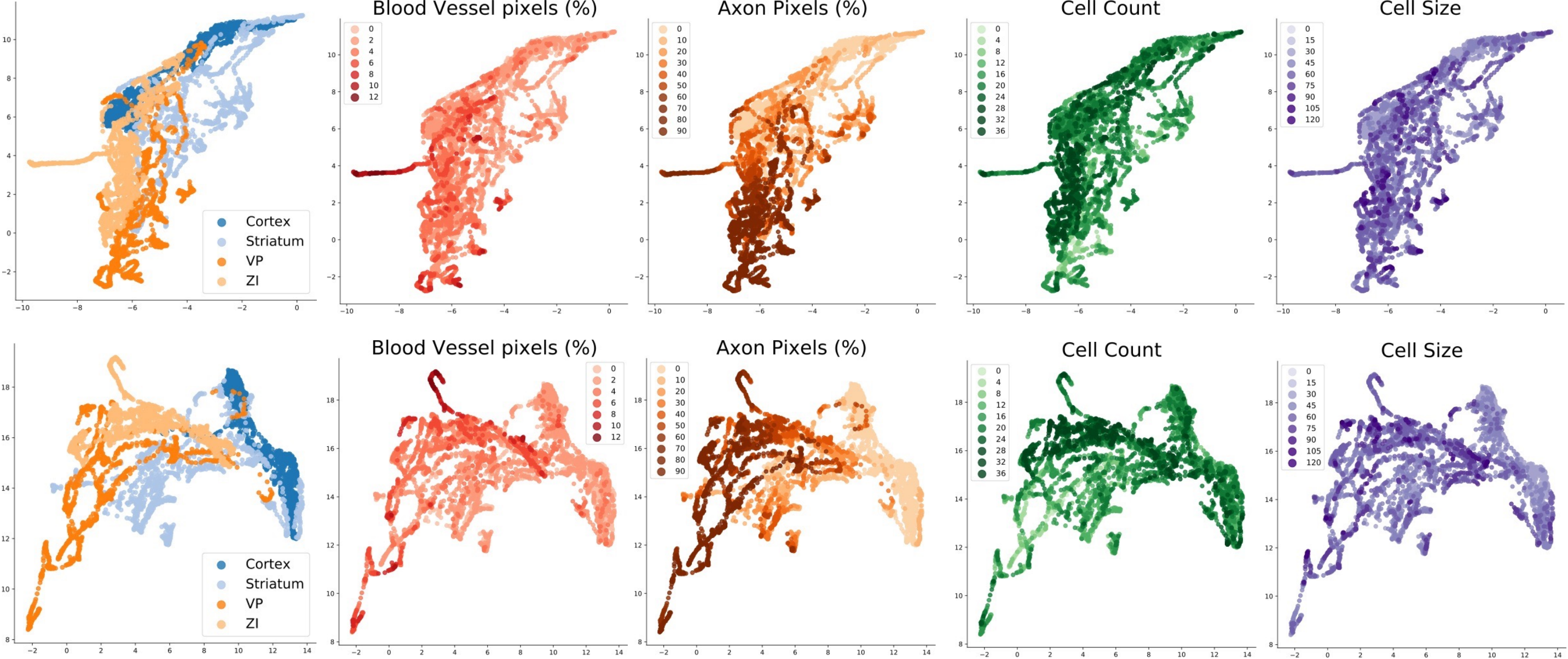}
\caption{\footnotesize {\em Visualizations of semantic features overlaid on two dimensional learned representations. } Here, the first row shows 2D U-map projections of learned BYOL (ROI-C1) embeddings and overlay the different semantic attributes on the latents . From left to right, we color the embeddings by brain area (class), \% blood vessels, \% axons, cell count, and cell size.
The second row shows the 2D projections  of learned BYOL  (ROI-C3) embedding which too are overlaid with different semantic attributes.
\vspace{-3mm}}
\label{fig:applatent}
\end{figure*}

Once the semantic features for different tasks are calculated, we extract the representations of all samples across the 4 cubes using the models trained in task 1 ROI-C1 (Table \ref{tab:task3}, top) and ROI-C3 (Table \ref{tab:task3}, bottom), and use scikit-learn to fit a linear regression in order to predict the semantic features and report the R$^2$ on the entire subvolume of interest. Furthermore, we provide visualizations of the representations and the corresponding semantic features of the 4 interest cubes using a BYOL encoder trained under two different dataset conditions (ROI-C1 and ROI-C3) in Figure \ref{fig:applatent}.

\label{app:task3}

%%%%%%%%%%  DATA ACCESS  %%%%%%%%%% 
\edit{
\section{Maintenance, Licensing, and Ethical Concerns}
The dataset, data access API, and dataloaders are maintained by the BossDB team (bossdb.org), which is tasked as the BRAIN Initiative archive of record for nanoscale connectomics datasets. The team is developing standards in accordance with FAIR principles (\url{https://www.go-fair.org/fair-principles/}) to ensure permanent identifiers and broad accessibility. The data are available under the CC-BY-4.0 license. The baseline code is available open source hosted in a github organization, and community forks and pull requests will be welcome, to be reviewed by the repository maintainers. As improvements are made to the baseline codebase for future challenges, changes will be pushed to the MTNeuro repository as a new versioned release.

The dataset used includes animal data which was collected under an approved IACUC protocol, as detailed in the original paper. There is no human derived data in this dataset. We emphasize this dataset is for development of algorithms for fundamental analysis of structural neuroscience data, but it is possible future efforts could use these data inappropriately for the development of clinically relevant algorithms which could result in negative societal impacts. This use is discouraged due to the unknown generalizations of high-resolution X-ray Microtomography to clinical modalities.

}

\end{document}